\documentclass[journal]{IEEEtran}
\pdfoutput=1
\IEEEoverridecommandlockouts
\usepackage{color}
\usepackage[x11names]{xcolor}
\usepackage{tikz}
\usetikzlibrary{shapes,arrows,arrows.meta}
\usepackage{mathtools}
\usepackage{amssymb}
\usepackage{lipsum}
\usepackage{float}
\usepackage{makecell}
\usepackage{longtable}
\usepackage{stfloats}
\usepackage{multirow}
\usepackage{amsmath}
\usepackage{bm}
\usepackage{algorithm}
\usepackage{algpseudocode}
\usepackage{booktabs}
\usepackage[numbers,sort&compress]{natbib}
\usepackage{natbib}
\usepackage{setspace}
\usepackage{balance}
\usepackage{soul}
\algdef{SE}[SUBALG]{Indent}{EndIndent}{}{\algorithmicend\ }%
\algtext*{Indent}
\algtext*{EndIndent}

\usepackage{xcolor}
\usepackage{threeparttable}

\allowdisplaybreaks
\usepackage{graphicx}
\usepackage{caption}
\usepackage{subfigure}
\usepackage[colorlinks,citecolor=blue,urlcolor=blue,bookmarks=false,hypertexnames=true]{hyperref}

\begin{document}

\title{\textsc{Doracamom}: Joint 3D Detection and Occupancy Prediction with Multi-view 4D Radars and Cameras for Omnidirectional Perception}

\author{\IEEEauthorblockA{
Lianqing Zheng\IEEEauthorrefmark{1},
Jianan Liu\IEEEauthorrefmark{1},
Runwei Guan,
Long Yang,
Shouyi Lu,
Yuanzhe Li,\\
Xiaokai Bai,
Jie Bai,
Zhixiong Ma\IEEEauthorrefmark{2},
Hui-Liang Shen,
and Xichan Zhu
}
\vspace{-5 mm}
\thanks{This work has been submitted to the IEEE for possible publication. Copyright may be transferred without notice, after which this version may no longer be accessible.}
\thanks{\IEEEauthorrefmark{1}Both authors contribute equally to the work and are co-first authors.}
\thanks{\IEEEauthorrefmark{2}Corresponding author.}

\thanks{Lianqing Zheng, Long Yang, Shouyi Lu, Zhixiong~Ma and Xichan Zhu are with the School of Automotive Studies, Tongji University, Shanghai, China. Email: \{zhenglianqing, yanglong, 2210803, mzx1978, zhuxichan\}@tongji.edu.cn.}
\thanks{Jianan Liu is with Momoni AI, Gothenburg, Sweden. Email: jianan.liu@momoniai.org.}
\thanks{Runwei Guan is with the Department of Computer Science and Engineering, The Hong Kong University of Science and Technology, Guangzhou, China. Email: runwei.guan@liverpool.ac.uk.}
\thanks{Yuanzhe Li is with the Chair of Automotive Engineering, Technische Universität Berlin, Berlin, Germany. Email: yuanzhe.li@campus.tu-berlin.de}
\thanks{Xiaokai Bai and Hui-Liang Shen are with College of Information Science and Electronic Engineering, Zhejiang University, Hangzhou 310027, China. Email: \{shawnnnkb, shenhl\}@zju.edu.cn}
\thanks{Jie Bai is with the School of Information and Electrical Engineering, Hangzhou City University, Hangzhou 310015, China. E-mail: baij@zucc.edu.cn.}
}


\maketitle

\begin{abstract}
3D object detection and occupancy prediction are critical tasks in autonomous driving, attracting significant attention. Despite the potential of recent vision-based methods, they encounter challenges under adverse conditions. Thus, integrating cameras with next-generation 4D imaging radar to achieve unified multi-task perception is highly significant, though research in this domain remains limited. In this paper, we propose Doracamom, the first framework that fuses multi-view cameras and 4D radar for joint 3D object detection and semantic occupancy prediction, enabling comprehensive environmental perception. Specifically, we introduce a novel \texttt{Coarse Voxel Queries Generator} that integrates geometric priors from 4D radar with semantic features from images to initialize voxel queries, establishing a robust foundation for subsequent Transformer-based refinement. To leverage temporal information, we design a \texttt{Dual-Branch Temporal Encoder} that processes multi-modal temporal features in parallel across BEV and voxel spaces, enabling comprehensive spatio-temporal representation learning. Furthermore, we propose a \texttt{Cross-Modal BEV-Voxel Fusion} module that adaptively fuses complementary features through attention mechanisms while employing auxiliary tasks to enhance feature quality. Extensive experiments on the OmniHD-Scenes, View-of-Delft (VoD), and TJ4DRadSet datasets demonstrate that Doracamom achieves state-of-the-art performance in both tasks, establishing new benchmarks for multi-modal 3D perception. Code and models can be available at \url{https://github.com/TJRadarLab/Doracamom}.
\end{abstract}


\section{Introduction}
\label{Introduction}

\begin{figure}[t]
    \centering
    \includegraphics[width=\linewidth]{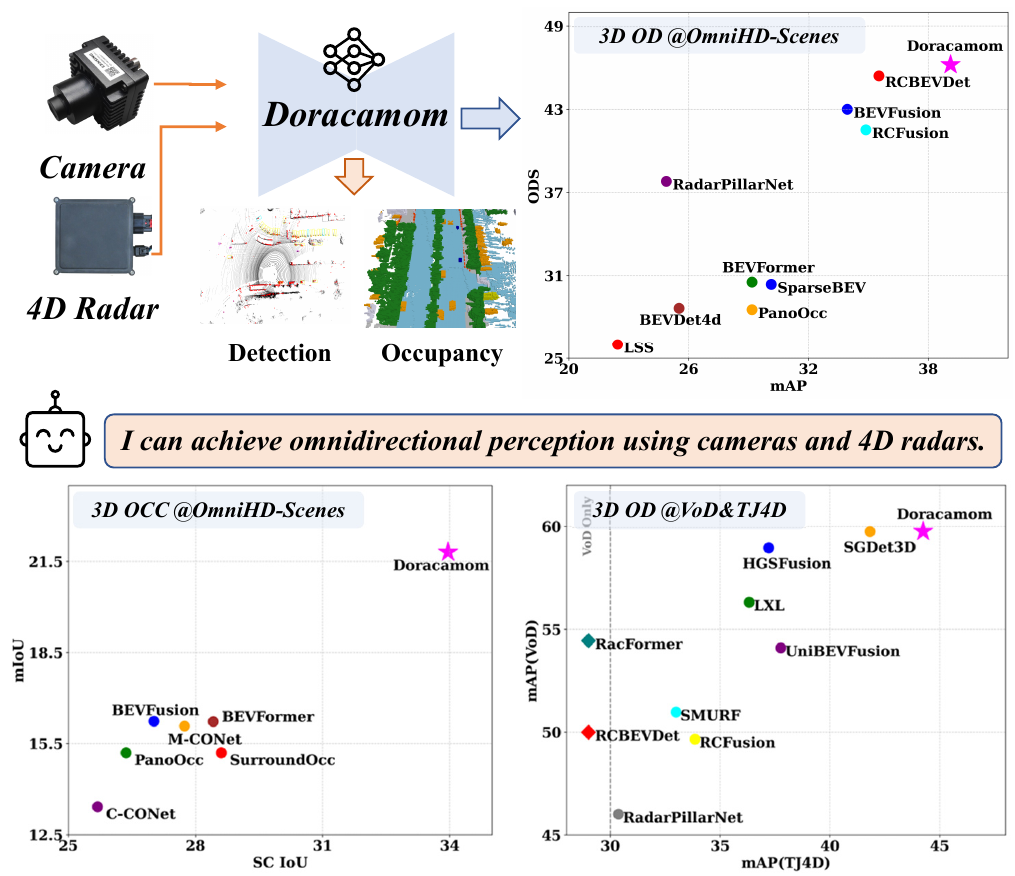}
    \vspace{-0.6cm}
    \caption{Performance comparison of different methods on 3D perception benchmarks. By effectively fusing camera and 4D radar inputs, Doracamom (marked with \textcolor{magenta}{\textbf{$\star$}}) consistently achieves superior performance across multiple metrics, outperforming existing camera-only and camera-radar fusion approaches in both 3D object detection and occupancy prediction tasks.}
    \label{fig:performance}
    \vspace{0.2cm}
\end{figure} 

\IEEEPARstart{A}{utonomous} driving technology is at the forefront of modern transportation revolution, attracting considerable attention. Autonomous driving systems typically include components such as environmental perception, trajectory prediction, and planning control to achieve self-driving capabilities. Accurate 3D perception is a critical foundation, focusing mainly on 3D object detection and semantic occupancy prediction tasks. 3D object detection employs 3D bounding boxes to locate foreground objects in the scene and predicts attributes such as category and velocity, and belongs to  sparse scene representation \cite{MM3DOD_Survey_TIV_2023}. In contrast, semantic occupancy uses fine-grained voxel representation to capture the geometric and semantic features of a scene, which is a form of dense scene representation \cite{xu2025occsurvey}. To accomplish these tasks, sensors such as onboard cameras, LiDAR, and millimeter-wave radars are commonly used to collect environmental data as input.

Among these sensors, LiDAR provides high-precision geometric information through dense point clouds \cite{Pointpillars, Centerpoint, C_L_OD_a, C_L_OD_b, PC3T, LEGO}, but its high cost and weather sensitivity limit deployment \cite{RaLiBEV}. Cameras offer rich visual information but lack depth perception and weather robustness \cite{ma20223d}. In contrast, radar provides reliable ranging and velocity measurements in adverse conditions. The all-weather reliability makes it as a key sensor in different fields, like autonomous driving, remote sensing, and SAR images \cite{radar_camera_survey,zhang2019high,zhang2019depthwise}. The 4D imaging radar, advancing beyond conventional $2+1$D radar, provides additional elevation information and higher resolution point clouds \cite{4DRADARSURVEY,10477463}. Therefore, combining 4D radar and cameras brings together the best of both, the all-weather reliability of radar and the rich semantic and texture details from cameras. Compared to expensive LiDAR systems, this fusion scheme offers a more attractive trade-off between performance, robustness, and cost.

Although fusing 4D radar and camera data for perception is a promising research area, existing methods still face several critical challenges. These challenges stem from the heterogeneity between the two modalities, i.e., 4D radar provides sparse coarse-grained geometric priors while cameras provide dense fine-grained semantic features. First, the sparsity of the radar makes query-based methods inefficient as they fail to leverage the valuable geometric priors from the radar. Second, it is a core challenge to effectively fuse these two highly heterogeneous data sources to meet the different feature requirements of multi-task perception. Finally, this combination differs from dense LiDAR-camera fusion, where an ``early fusion" before temporal modeling is typically utilized; for sparse 4D radar, this approach is sub-optimal, making it extremely difficult to capture and preserve the unique temporal dynamics of each modality.

To this end, we introduce Doracamom, a unified framework that, for the first time, fuses multi-view cameras with 4D radar point clouds to handle 3D object detection and semantic occupancy prediction tasks simultaneously. The main contributions of this paper are concluded as below:

\begin{itemize}
\item We propose Doracamom, which is the first unified framework to fuse cameras and 4D radar for joint 3D object detection and occupancy prediction, enabling comprehensive environmental perception and understanding. 

\item Specifically, we design three key components to enhance model performance. The coarse voxel query generator (CVQG) leverages 4D radar geometric cues and image semantic information to establish well-initialized voxel queries for effective feature refinement. Dual-branch temporal encoder (DTE) enables parallel temporal modeling in both BEV and voxel domains, capturing comprehensive spatio-temporal representations of the scene. A cross-modal BEV-voxel fusion (CMF) module adaptively fuses complementary features through attention mechanisms, incorporating auxiliary binary occupancy prediction and BEV segmentation tasks to guide the feature learning process for more discriminative representations.
\item The comprehensive experiments show that Doracamom achieves state-of-the-art performance and makes new benchmark for both 3D object detection and occupancy prediction with 4D radar and camera fusion, on several 4D radar datasets including OmniHD-Scenes \cite{OmniHDScenes}, VoD \cite{palffy2022multi}, and TJ4DRadSet \cite{zheng2022tj4dradset}, as shown in Fig. \ref{fig:performance}.

\end{itemize}

\section{Related Works}
\label{Related_Works}

\subsection{3D Perception with Camera}

Recent 3D perception research has primarily focused on vision-based approaches. While early work directly regressed 3D attributes from single images \cite{wang2021fcos3d}, such monocular methods suffer from occlusion and viewpoint changes. This has led to increased interest in multi-view approaches that transform 2D features into 3D space \cite{bevsurvey,lss,huang2021bevdet,bevdepth,bevformer} using BEV representations.


Existing methods can be categorized into three main approaches. The first uses depth prediction, starting with LSS \cite{lss} which lifts 2D features to BEV space through depth distributions and voxel pooling. This was extended by BEVDepth \cite{bevdepth}, BEVDet \cite{huang2021bevdet}, and BEVDet4D \cite{huang2022bevdet4d} with LiDAR supervision and temporal fusion \cite{bevpoolv2}. The second approach employs back projection, where methods like OFTNet \cite{Roddick2018OrthographicFT} and simpleBEV \cite{harley2023simplebev} project 3D voxels onto 2D images for feature sampling. The third approach is attention-based, where BEVFormer \cite{bevformer} uses BEV queries and deformable attention \cite{deformabledetr} for adaptive feature aggregation, while DETR3D \cite{wang2022detr3d}, PETR \cite{liu2022petr}, and Sparse4D \cite{lin2022sparse4d} demonstrate strong performance using object queries without explicit BEV features.


Vision-based occupancy prediction transforms 2D features to 3D space to obtain BEV features \cite{yu2023flashocc}, TPV features \cite{TPVFormer}, or voxel features \cite{surroundocc}. Methods are categorized into projection-based, depth-based, and cross-attention-based approaches \cite{xu2025occsurvey}. MonoScene \cite{monoscene} uses projection and U-Net for semantic completion, while FlashOcc \cite{yu2023flashocc} and FastOcc \cite{fastocc} follow LSS \cite{lss} to predict depth distributions, using channel-to-height operations \cite{yu2023flashocc} for memory efficiency. SurroundOcc \cite{surroundocc}, PanoOcc \cite{wang2024panoocc}, and TPVFormer \cite{TPVFormer} employ deformable attention \cite{deformabledetr} for feature aggregation. Recent works include AdaptiveOcc \cite{AdaptiveOcc}, using octree for voxel representation, and LinkOcc \cite{LinkOcc}, incorporating sparse queries and temporal association through near-online training and contrastive learning.

\subsection{3D Perception with Traditional Radar and Camera Fusion}

Conventional automotive radar is constrained by its angular resolution, resulting in sparse point clouds lacking height information, necessitating camera-radar fusion for enhanced perception \cite{radar_camera_survey}. CenterFusion \cite{nabati2021centerfusion} generates frustum ROIs using 2D detectors and associates them with radar pillars. Feature-level fusion methods work at either BEV level \cite{kim2023crn,lin2024rcbevdet} or proposal stage \cite{kim2023craft,pang2023transcar}. CRN \cite{kim2023crn} uses radar for BEV transformation and multi-modal deformable attention for feature alignment. RCBEV \cite{zhou2023bridging} combines point fusion and ROI fusion, while RCBEVDet \cite{lin2024rcbevdet} employs RadarBEVNet with dual representations and RCS-aware BEV features, employing a cross-modal attention mech
anism to merge radar and image BEV features. CRAFT \cite{kim2023craft} associates proposals with radar points in polar coordinates, while CramNet \cite{hwang2022cramnet} uses ray-constrained attention for geometric correspondence. TransCAR \cite{pang2023transcar} and FUTR3D \cite{futr3d}, both extending DETR3D \cite{wang2022detr3d}, utilize object queries for radar-camera feature interaction. Notably, RaCFormer \cite{chu2025racformer} further enhances this query-based paradigm by introducing a radar-guided depth estimation to improve the view transformation and adaptively samples features from both PV and BEV views. CRT-Fusion \cite{CRTFusion} introduces a temporal fusion framework that estimates object motion to guide the alignment and recurrent fusion of multi-timestamp BEV features.


Beyond 3D detection, recent works explore camera-radar fusion for occupancy prediction. Occfusion \cite{occfusion} employs dynamic 2D/3D fusion for multi-level feature representation, while LiCROcc \cite{ma2024licrocc} uses cross-modal distillation modules for semantic scene completion. However, the sparsity and lack of height information in traditional radar remain challenging for 3D semantic occupancy tasks.

\subsection{Recent Progress on Perception with 4D Radar}

The 4D imaging radar offers significant improvements over conventional radar in resolution and point cloud density, along with elevation resolution, greatly expanding its potential applications \cite{4DRADARSURVEY,10477463}. The recent emergence of 4D radar datasets \cite{meyer2019astyx,zheng2022tj4dradset,palffy2022multi,paek2022kradar,zhang2023ntu4dradlm,yao2024waterscenes,OmniHDScenes} has spurred research in this field. Some studies focus on leveraging the characteristics of 4D radar for multi-object tracking \cite{pan2024ratrack,liu2024framework}. Other researches utilize 4D radar point clouds for mapping and localization \cite{zhuang20234diriom,ronet,4DRVONet}, scene flow estimation \cite{ding2023hidden,SceneFlow1}, panoptic segmentation \cite{Achelous,ASYVRNet} and visual grounding \cite{WaterVG,NanoMVG,Talk2Radar}, among other applications.

In 3D object detection, many studies have achieved remarkable results using 4D radar and multi-sensor fusion. RPFA-Net \cite{xu2021rpfa} and SMURF \cite{liu2023smurf} focus on radar feature extraction, while InterFusion \cite{wang2022interfusion} employs a self-attention mechanism to learn features from both LiDAR and 4D radar modalities. RCFusion \cite{zheng2023rcfusion} leverages RadarPillarNet for hierarchical feature extraction from 4D radar point clouds and effectively fuses BEV features using interactive attention. Recent works have introduced more advanced fusion strategies. UniBEVFusion \cite{zhao2024unibevfusion} employs RDL module for depth estimation and UFF for multi-modal feature integration. LXL \cite{xiong2023lxl} combines predicted depth with radar occupancy grids to enhance image view transformation. SGDet3D \cite{SGDet3D} utilizes dual-branch fusion and object-oriented attention for feature interaction, while HGSFusion \cite{HGSFusion} leverages hybrid point generation and dual synchronization. DPFT \cite{DPFT} achieves effective fusion through 3D space query sampling.

For occupancy prediction, RadarOcc \cite{ding2024radarocc} performs binary occupancy prediction using 4D radar tensors, but its resource-intensive processing limits real-world applications. While radar point clouds are widely used in autonomous driving systems, research on 3D semantic occupancy prediction with 4D radar point clouds remains limited.

\section{Methodology}

\subsection{Overall Architecture}
\begin{figure*}[t]
    \centering
    \includegraphics[width=\linewidth]{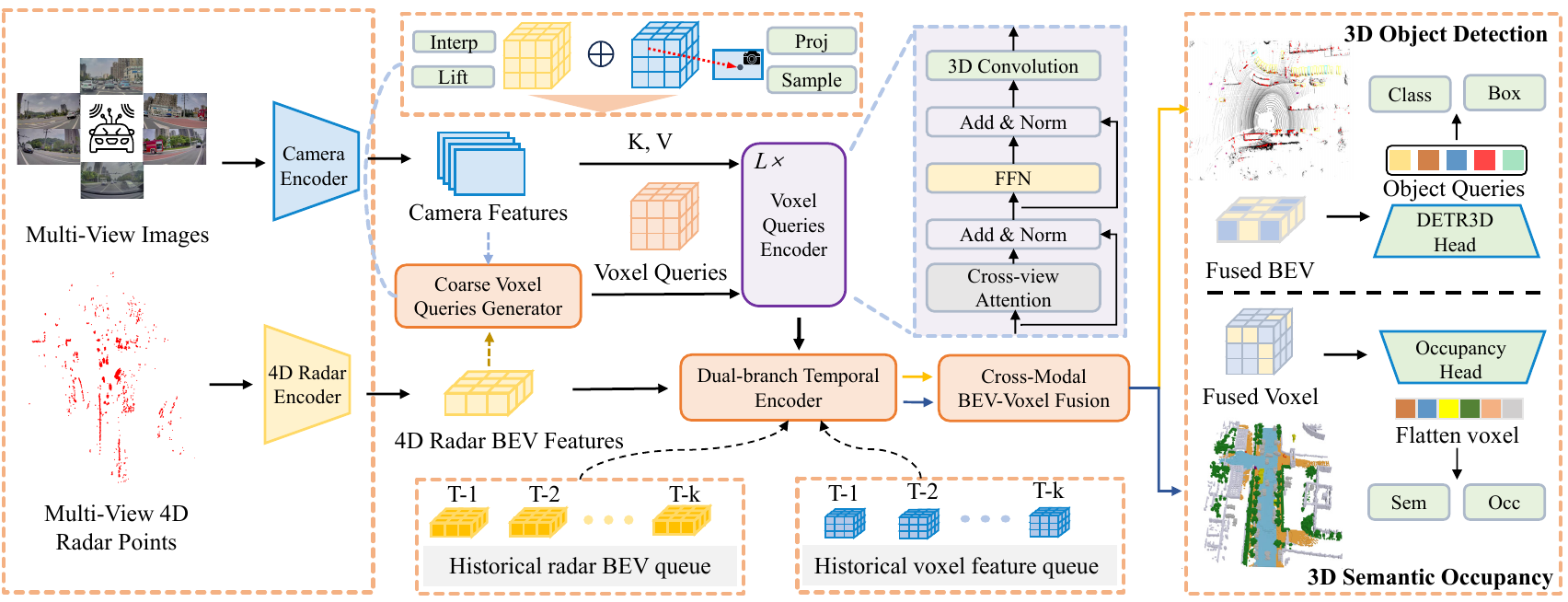}
    \vspace{-0.6cm}
    \caption{\textbf{The overall framework of Doracamom.} Initially, the camera encoder and 4D radar encoder are utilized to extract multi-camera features and radar BEV features, respectively. Subsequently, the \textbf{coarse voxel query generator} employs geometric priors derived from 4D radar features and semantic priors from image features to generate coarse voxel queries. These queries are then refined by a stacked voxel queries encoder to obtain local fine-grained voxel features. A \textbf{dual-branch temporal encoder} is employed to fuse historical BEV and voxel features with current frame features, leveraging temporal clues. The output radar BEV and image voxel features are fed into the \textbf{cross-modal BEV-voxel fusion} module for adaptive fusion, resulting in the final BEV and voxel representations. Finally, the obtained representations are used to predict 3D detection and semantic occupancy for the current scene.}
    \label{fig:model}
    \vspace{0.2cm}
\end{figure*}

We present \textsc{Doracamom}, a unified multi-task framework that fuses multi-view images and 4D radar point clouds for joint 3D object detection and occupancy prediction. The overall architecture is shown in Fig. \ref{fig:model}. 

Initially, multi-view images and 4D radar point clouds are fed into the camera and 4D radar encoders to extract image 2D features and 4D radar BEV features, respectively. These features are then passed to the coarse voxel query generator, which combines image and radar features to generate geometrically-semantic-aware coarse-grained voxel queries. The voxel query encoder iteratively enhances fine-grained voxel features through stacked transformer blocks using cross-view attention. Subsequently, a dual-branch temporal encoder leverages temporal cues to enhance both BEV and voxel feature representations. The cross-modal BEV-voxel fusion module adaptively integrates the feature representations from both modalities, obtaining the final BEV and voxel features, which are then fed into the multi-task head for prediction.


\subsection{Camera \& 4D Radar Encoders}

In the feature extraction stage, we employ a decoupled architecture for independent high-dimensional feature extraction from two input modalities. The camera encoder processes multi-view images represented as $I \in \mathbb{R}^{N_C \times 3 \times H_I \times W_I}$, where $N_C$ is the number of cameras, $H_I$ and $W_I$ are the image height and width, and 3 corresponds to the RGB channels. Feature extraction is performed using a shared ResNet-50 \cite{he2016resnet} backbone network and a Feature Pyramid Network (FPN) \cite{lin2017fpn} as the neck structure, which obtains multi-scale features and simplifies them into a single-scale representation $\mathcal{F}_{I} = \{F_I^i\}_{i=1}^{N_C}$, where $F_I^i\in \mathbb{R}^{C \times H_C \times W_C}$ is the feature of the $i$-th view, $C$ denotes the feature dimension, and $H_C$ and $W_C$ are the spatial dimensions of the feature map.

The 4D radar encoder processes input point clouds $\mathcal{P} \in \mathbb{R}^{N_R \times D}$, where $N_R$ and $D$ represent the number of points and attribute dimension, respectively. Various learning-based methods \cite{zhang2020hyperli,zhang2021balance,xu2021rpfa} exist for extracting features from radar data. To balance both efficiency and performance, We adopt RadarPillarNet \cite{zheng2023rcfusion}, which generates pseudo images through hierarchical feature extraction. The encoded features are then processed by SECOND and SECONDFPN \cite{yan2018second} to produce refined 4D radar BEV features $\mathcal{F}_{R} \in \mathbb{R}^{C_R \times H \times W}$, with $C_R$ representing the feature dimension and $(H,W)$ denoting the BEV resolution.

\subsection{Coarse Voxel Queries Generator}

\begin{figure}[t]
    \centering
    \includegraphics[width=\linewidth]{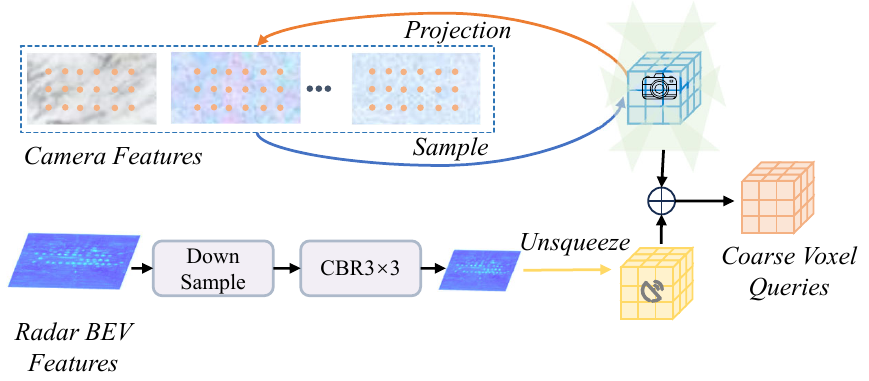}
    \vspace{-0.6cm}
    \caption{Illustration of the proposed \textbf{Coarse Voxel Queries Generator}. It  combines radar BEV features and multi-view image features to initialize voxel queries with both geometric and semantic priors.}
    \label{fig:coarse}
\end{figure} 
We formulate the 3D voxel queries in space as $\mathcal{Q}\in\mathbb{R}^{ C \times H_V\times W_V\times Z_V}$, where $(H_V,W_V,Z_V)$ denotes the voxel grid resolution.  To reduce computational overhead, we set the BEV plane resolution of the voxel grid as $(H_V,W_V)=(\frac{H}{2},\frac{W}{2})$. While existing approaches \cite{surroundocc, wang2024panoocc} conventionally utilize random initialization for voxel query generation, this methodology can hinder training efficiency and limit final performance, as the model must learn scene representations without any initial guidance. To address this limitation and enhance the fidelity of view transformation, we introduce a novel initialization method that integrates geometric priors derived from 4D radar data with semantic features extracted from images. This integration enables the generation of coarse-grained voxel queries with both geometric and semantic priors, establishing a more robust foundation for subsequent refinement procedures.  Inspired by \cite{harley2023simplebev, fastocc}, we design a voxel query initialization pipeline as followings. 

In the radar feature processing phase, we initially transform the radar BEV features $\mathcal{F}_R$ through \textit{Bilinear Interpolation} to align with the voxel grid, yielding $\mathcal{F}_{R}^{'}\in\mathbb{R}^{C_R \times H_V\times W_V}$. Subsequently, we further optimize the feature channels using a Conv-BN-ReLU (CBR). By applying a simple \textit{"unsqueeze"} operation to expand the 2D BEV features along the height dimension, we obtain the radar 3D voxel features $\mathcal{Q}_R$, which can be mathematically expressed as:

\vspace{-5pt}
\begin{equation}
\mathcal{Q}_R = \mathtt{Unsqueeze}(\mathtt{CBR}(\mathcal{F}_{R}^{'}), Z_V)
\end{equation}

For image feature processing, we adopt a methodology similar to \cite{harley2023simplebev, li2024fastbev}. We first define 3D reference points $\mathcal{P}_{ref}\in\mathbb{R}^{H_V\times W_V\times Z_V\times 3}$ within the ego-vehicle coordinate frame, based on the shape of the 3D voxel queries. Concurrently, we initialize the voxel features $\mathcal{Q}_I\in\mathbb{R}^{C\times H_V\times W_V\times Z_V}$ to zero. The transformation matrix $\mathbf{T}_{e\to I}\in\mathbb{R}^{3\times 4}$ from the ego-vehicle coordinate frame to the image pixel coordinate is then computed using the camera's intrinsic matrix $K\in\mathbb{R}^{3\times 4}$ and extrinsic matrix $\mathbf{T}_{e\to c}\in\mathbb{R}^{4\times 4}$:

\vspace{-5pt}
\begin{equation}
\mathbf{T}_{e\to I} = K\mathbf{T}_{e\to c}
\end{equation}

Leveraging $\mathbf{T}_{e\to I}$, we project reference points onto the each image planes to obtain their corresponding coordinates $(x,y,z)$ on the feature maps. Valid points are determined by two criteria: $(x,y)$ must lie within the feature map boundaries and $z$ must be positive. The feature sampling process employs nearest neighbor interpolation, with a \textit{"last-update"} strategy resolving overlapping multi-view regions. The final coarse-grained voxel queries are obtained by adding $\mathcal{Q}_R$ and $\mathcal{Q}_I$.

\subsection{Voxel Queries Encoder}

To enhance and refine voxel queries, we employ an $L$-layer Transformer-based architecture for feature encoding. Inspired by \cite{deformabledetr,bevformer,surroundocc}, we adopt deformable attention for cross-view feature aggregation, which not only alleviates occlusion and ambiguity issues but also improves efficiency by reducing training time.

In the cross-view attention module, the inputs consist of voxel queries $\mathcal{Q}\in\mathbb{R}^{C\times H_V\times W_V\times Z_V}$, corresponding 3D reference points $\mathcal{P}_{ref}\in\mathbb{R}^{H_V\times W_V\times Z_V\times 3}$, and image features $\mathcal{F}_{I} = \{F_I^i\}_{i=1}^{N_C}$. The 3D reference points are projected into 2D views using camera parameters, and image features are sampled and weighted from the hit views. The output features $\mathbf{Q}_O$ can be expressed as:

\vspace{-5pt}
\begin{equation}
\mathbf{Q}_O^{P} = \frac{1}{|\mathcal{V}_{\text{hit}}|} \sum_{i \in \mathcal{V}_{\text{hit}}} \mathtt{DeformAttn}(\mathbf{Q}^p, \mathtt{Proj}(\mathcal{P}_{ref}^{p} \ , \ i), F_I^i)
\end{equation}
where $i$ denotes the image view index, $\mathbf{Q}^p$ and $\mathbf{Q}_O^{P}$ represent the $p$-th voxel feature and its output feature respectively, $\mathtt{Proj}(\mathcal{P}_{ref}^{p}, i)$ denotes the projection function that maps 3D reference points to the $i$-th image view, and $\mathcal{V}_{\text{hit}}$ represents the set of visible views. Similar to \cite{surroundocc}, we employ 3D convolutions for neighboring voxel feature interaction instead of computationally expensive 3D self-attention.
\subsection{Dual-branch Temporal Encoder}

Temporal information plays a crucial role in perception systems. Existing methods \cite{bevformer,li2024fastbev, wu2023leveraging,wu2024temporal}, have demonstrated that leveraging temporal features can effectively address occlusion issues, enhance scene understanding, and improve the accuracy of motion state estimation. However, these approaches are limited to temporal modeling in a single feature space, making it challenging to to capture comprehensive spatio-temporal representations. 

\begin{figure}[t]
    \centering
    \includegraphics[width=\linewidth]{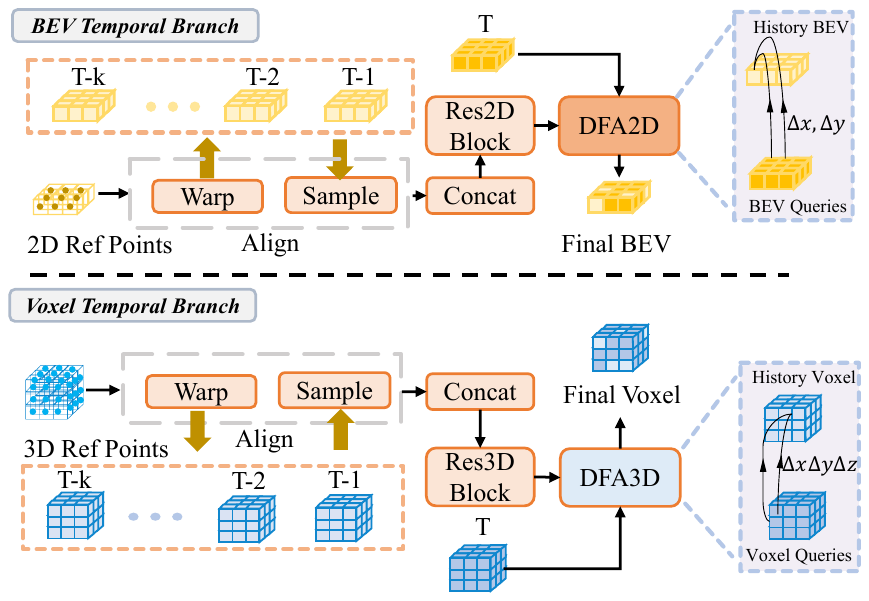}
    \vspace{-0.6cm}
    \caption{Illustration of the proposed \textbf{Dual-branch Temporal Encoder}. To eliminate misalignment caused by ego motion, ego poses are used to warp the 2D and 3D reference points and sample features to the current frame. Historical features are then merged using ResNet2D/3D blocks to reduce cross-frame feature interaction and enhance efficiency. To mitigate the impact of moving objects, 2D and 3D deformable attention mechanisms are employed to adaptively fuse features from the current and historical frames.}
    \label{fig:temporal}
\end{figure}

To address this limitation, we propose a novel Dual-branch Temporal Encoder module that processes multi-modal temporal features in parallel across BEV and voxel spaces, as shown in Fig. \ref{fig:temporal}. Specifically, the radar BEV branch excels at capturing global geometric features while the image voxel branch focuses on preserving fine-grained semantic information. This complementary dual-branch design not only provides diverse representational capabilities in feature expression and temporal modeling but also achieves an optimized balance between computational cost and feature expressiveness. Additionally, the feature redundancy mechanism significantly enhances the robustness of the perception system.

In temporal feature fusion, a key challenge is feature misalignment caused by ego-motion and dynamic object movement. To address the feature displacement induced by ego-motion, we propose a pose transformation-based feature alignment strategy that precisely aligns historical features with the current frame. Specifically, in the voxel temporal branch, given 3D reference points $\mathcal{P}_{t}^{3D} \in\mathbb{R}^{H_V\times W_V\times Z_V\times 3}$ at current frame $t$ in the ego-vehicle coordinate system, we utilize pose sequences from both current and historical frames $\{\mathbf{T}_t, \mathbf{T}_{t-1}, \mathbf{T}_{t-2}, ..., \mathbf{T}_{t-k}| \mathbf{T}_{t-i} \in \mathbb{R}^{4\times4}\}$ to warp these reference points to corresponding historical timestamps. Subsequently, we employ trilinear interpolation to sample the temporally aligned features. The detailed computation process is as follows:

\vspace{-5pt}
\begin{equation}
\mathcal{P}_{t-i}^{3D} = \mathbf{T}_{t-i}^{-1} \cdot \mathbf{T}_{t} \cdot \mathcal{P}_{t}^{3D}
\end{equation}

\vspace{-10pt}

\begin{equation}
\mathbf{F}_{(t-i) \rightarrow t}^{vox} = \mathtt{GridSample3D}(\mathbf{F}_{t-i}^{vox}, \mathcal{P}_{t-i}^{3D})
\end{equation}
where $\mathcal{P}_{t-i}^{3D}$ represents the transformed 3D reference points at historical timestamp $t-i$, and $\mathbf{F}_{(t-i) \rightarrow t}^{vox} \in\mathbb{R}^{C\times H_V\times W_V\times Z_V}$ denotes the aligned historical features, forming a temporal feature sequence $\{\mathbf{F}_{(t-1) \rightarrow t}^{vox}, \mathbf{F}_{(t-2) \rightarrow t}^{vox}, ..., \mathbf{F}_{(t-k) \rightarrow t}^{vox}\}$. Similarly, for the BEV temporal branch, we define 2D reference points $\mathcal{P}_{t}^{2D} \in\mathbb{R}^{H\times W\times 3}$ by setting the height dimension to zero. The historical features are obtained through bilinear interpolation after warping these reference points to corresponding timestamps, which can be formulated as:

\vspace{-5pt}
\begin{equation}
\mathcal{P}_{t-i}^{2D} = \mathbf{T}_{t-i}^{-1} \cdot \mathbf{T}_{t} \cdot \mathcal{P}_{t}^{2D}
\end{equation}

\vspace{-10pt}

\begin{equation}
\mathbf{F}_{(t-i) \rightarrow t}^{BEV} = \mathtt{GridSample2D}(\mathbf{F}_{t-i}^{BEV}, \mathcal{P}_{t-i}^{2D})
\end{equation}
where $\mathcal{P}_{t-i}^{2D}$ represents the transformed 2D reference points at historical timestamp $t-i$, and $\mathbf{F}_{(t-i) \rightarrow t}^{BEV}$ denotes the aligned historical features, forming a temporal feature sequence ${\mathbf{F}_{(t-1) \rightarrow t}^{BEV}, \mathbf{F}_{(t-2) \rightarrow t}^{BEV}, ..., \mathbf{F}_{(t-k) \rightarrow t}^{BEV}}$ for the BEV branch.

To further mitigate feature misalignment caused by dynamic objects, we employ deformable attention to adaptively fuse features between current and historical frames. For the voxel temporal branch, we first concatenate the aligned historical features and process them through a simple Res3D block for efficient feature integration, which can be formulated as:

\vspace{-10pt}
\begin{equation}
\mathbf{F}_{hist}^{vox} = \mathtt{ResBlock3D}(\mathtt{Concat}({\mathbf{F}_{(t-1) \rightarrow t}^{vox}, \ ... \ , \mathbf{F}_{(t-k) \rightarrow t}^{vox}}))
\end{equation}

Subsequently, we employ deformable attention to adaptively integrate current and historical features. The fusion process can be formulated as:

\vspace{-10pt}
\begin{equation}
\mathbf{F}_{O}^{vox_p} = \sum_{V \in \{\mathbf{F}_{t}^{vox}, \mathbf{F}_{hist}^{vox}\}} \mathtt{DeformAttn3D}(\mathbf{F}^{vox_p}  , p^{3D} \ , V)
\end{equation}
where $\mathbf{F}^{vox_p}$ and $\mathbf{F}_{O}^{vox_p}$ denotes the voxel feature and its output feature located at $p^{3D} = (x, y, z)$, respectively. 

For the BEV temporal branch, a similar process is applied. The historical BEV features are first concatenated and processed through a Res2D block:

\vspace{-10pt}
\begin{equation}
\mathbf{F}_{hist}^{BEV} = \mathtt{ResBlock2D}(\mathtt{Concat}({\mathbf{F}_{(t-1) \rightarrow t}^{BEV}, \ ... \ , \mathbf{F}_{(t-k) \rightarrow t}^{BEV}}))
\end{equation}

Then, deformable attention is employed for feature fusion:

\vspace{-10pt}
\begin{equation}
\mathbf{F}_{O}^{BEV_p} = \sum_{V \in \{\mathbf{F}_{t}^{BEV}, \mathbf{F}_{hist}^{BEV}\}} \mathtt{DeformAttn2D}(\mathbf{F}^{BEV_p}, p^{2D}, V)
\end{equation}
where $\mathbf{F}^{BEV_p}$ and $\mathbf{F}_{O}^{BEV_p}$ denotes the BEV feature and its output feature located at $p^{2D} = (x, y)$, respectively.

This comprehensive approach ensures robust temporal feature fusion by combining global geometric patterns from the BEV branch and fine-grained semantic details from the voxel branch, leading to more accurate perception results.

\subsection{Cross-Modal BEV-Voxel Fusion Module}

\begin{figure}[t]
    \centering
    \includegraphics[width=\linewidth]{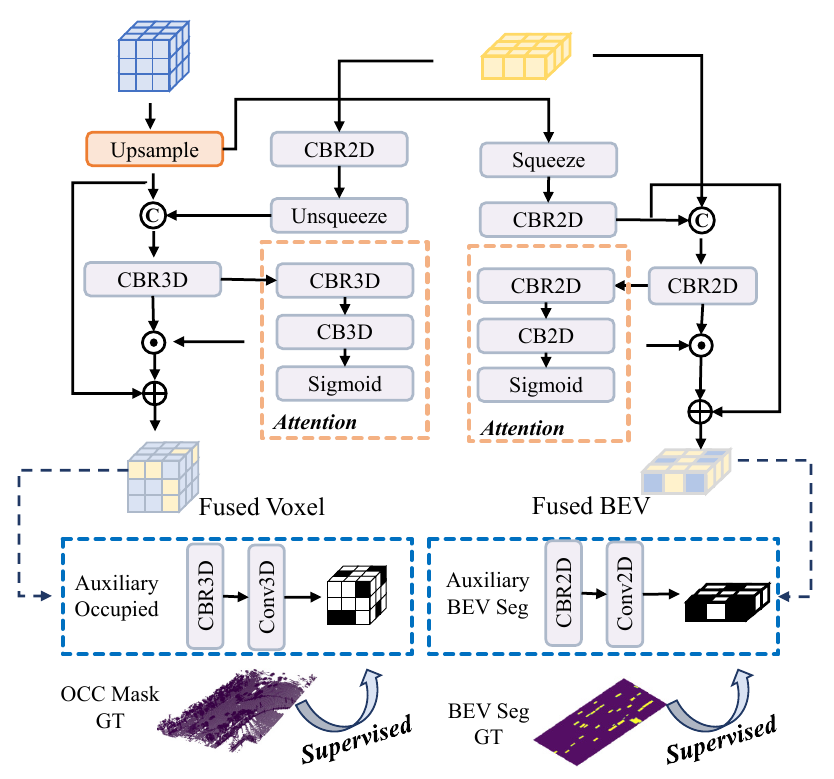}
    \vspace{-0.6cm}
    \caption{Illustration of the proposed \textbf{Cross-Modal BEV-Voxel Fusion} module. Complementary information from both modalities is adaptively fused to generate BEV and voxel features for downstream task decoding. Additionally, two auxiliary tasks which predict the occupied/non-occupied binary 3D occupancy probability and foreground/background BEV segmentation mask, are incorporated to enhance the quality of the generated 3D occupancy and BEV features.}
    \label{fig:fusion}
    \vspace{0.2cm}
\end{figure} 

To effectively leverage the temporally-enhanced features from both voxel and BEV spaces, we propose a Cross-Modal BEV-Voxel Fusion module that generates geometrically and semantically rich multi-modal representations for downstream multi-task decoding. As illustrated in Fig.~\ref{fig:fusion}, this module adaptively fuses heterogeneous features through attention-weighted mechanisms while employing auxiliary tasks to further enhance the quality of generated features.

Specifically, the module first upsamples the low-resolution voxel features through a 3D deconvolution block to obtain high-resolution features $\mathbf{F}^{vox} \in \mathbb{R}^{C \times H \times W \times Z}$ for subsequent fusion. For voxel feature enhancement, the radar BEV features $\mathbf{F}^{BEV}$ are first processed through Conv-BN-ReLU blocks in 2D to reshape the feature channels, followed by an unsqueeze operation that expands the 2D BEV features along the height dimension. The expanded features are then concatenated with voxel features and processed through convolutional blocks to reduce channel dimensions. Finally, a residual structure with attention mechanism is employed to obtain the fused features. This process can be formulated as:

\vspace{-10pt}
\begin{equation}
\mathbf{F}_{temp}^{vox} = f_{3D}(\mathtt{Concat}[\mathbf{F}^{vox}, \mathtt{Unsqueeze}(f_{2D}(\mathbf{F}^{BEV}))])
\end{equation}

\vspace{-10pt}

\begin{equation}
\mathbf{F}_{fused}^{vox} = \mathbf{F}^{vox} + \sigma(g_{3D}(\mathbf{F}_{temp}^{vox})) \odot \mathbf{F}_{temp}^{vox}
\end{equation}
where $f_{2D}(\cdot)$ and $f_{3D}(\cdot)$ represent the Conv-BN-ReLU blocks in 2D and 3D, $g_{3D}(\cdot)$ represents the convolutional operations in the attention block, $\sigma(x) = 1/(1+e^{-x})$ is the sigmoid function, and $\odot$ denotes element-wise multiplication. $\mathbf{F}_{temp}^{vox}$ and $\mathbf{F}_{fused}^{vox}$ are intermediate and final fused voxel features.

Similarly, for BEV feature enhancement, the voxel features $\mathbf{F}^{vox}$ are first compressed along the height dimension through a squeeze operation, followed by CBR2D blocks to adjust feature channels, obtaining transformed features $\mathbf{F}^{BEV'} \in \mathbb{R}^{C \times H \times W}$. The processed features are then concatenated with radar BEV features and refined through convolutional blocks. A similar residual attention structure is employed to obtain the final fused BEV features. This process can be formulated as:

\vspace{-10pt}
\begin{equation}
\mathbf{F}_{temp}^{BEV} = f_{2D}(\mathtt{Concat}[\mathbf{F}^{BEV}, f_{2D}(\mathtt{Squeeze}(\mathbf{F}^{vox}))])
\end{equation}

\vspace{-10pt}

\begin{equation}
\mathbf{F}_{fused}^{BEV} = \mathbf{F}^{BEV'} + \sigma(g_{2D}(\mathbf{F}_{temp}^{BEV})) \odot \mathbf{F}_{temp}^{BEV}
\end{equation}
where $g_{2D}(\cdot)$ represents the convolutional operations in the 2D attention block, and $\mathbf{F}_{temp}^{BEV}$ and $\mathbf{F}_{fused}^{BEV}$ are intermediate and final fused BEV features, respectively.

Additionally, to enhance the final feature representations and improve feature quality for subsequent decoding, we incorporate auxiliary tasks with explicit supervision. For the fused voxel features, we estimate binary occupancy masks $\mathbf{M}^{vox} \in \mathbb{R}^{H \times W \times Z}$ through 3D convolutional blocks, supervised by binary occupied/non-occupied ground truth derived from semantic occupancy labels. Similarly, for the fused BEV features, we project detection ground truth onto the BEV plane to obtain binary segmentation ground truth representing foreground objects and background, and generate corresponding prediction mask $\mathbf{M}^{BEV} \in \mathbb{R}^{H \times W}$ through 2D convolutional blocks. Both auxiliary tasks are supervised using a combination of Dice and Binary Cross-Entropy (BCE) losses. These auxiliary binary supervision signals help guiding the feature learning process, ensure the generation of more discriminative features progressively in the suitable area for downstream tasks with more specific semantic information, i.e., 3D semantic occupancy prediction and 3D object detection.

\subsection{Multi-task Training}
Based on the enhanced geometric and semantic-aware BEV and voxel representations, we can perform end-to-end training for joint 3D object detection and occupancy prediction, enabling comprehensive perception capabilities.


For 3D object detection, following \cite{bevformer}, we use DETR-based head unless specified otherwise. The head takes the fused BEV features $\mathbf{F}_{fused}^{BEV}$ as input and directly predicts object categories and attributes. The detection loss $\mathcal{L}_{det}$ consists of a focal loss for classification and an L1 loss for regression, which can be formulated as:

\vspace{-5pt}
\begin{equation}
\mathcal{L}_{det} = \lambda_{1}\mathcal{L}_{cls} + \lambda_{2}\mathcal{L}_{reg}
\end{equation}
where $\lambda_{1}=2.0$ and $\lambda_{2}=0.25$ are hyperparameters to balance the classification and regression losses.

For occupancy prediction, we employ a simple MLP on the fused voxel features $\mathbf{F}_{fused}^{vox}$ to predict semantic occupancy for each voxel. The occupancy prediction loss $\mathcal{L}_{occ}$ consists of three components: a primary cross-entropy loss $\mathcal{L}_{ce}$ for basic supervision, and two affinity losses $\mathcal{L}_{scal}^{geo}$ and $\mathcal{L}_{scal}^{sem}$ proposed by \cite{monoscene} to optimize scene-wise and class-wise metrics, respectively. The occupancy loss $\mathcal{L}_{occ}$ can be formulated as:

\vspace{-5pt}
\begin{equation}
\mathcal{L}_{occ} = \mathcal{L}_{ce} + \mathcal{L}_{scal}^{geo} + \mathcal{L}_{scal}^{sem}
\end{equation}

For both foreground/background BEV segmentation and occupied/non-occupied binary occupied prediction auxiliary tasks, we employ a combination of BCE loss and Dice loss for supervision. The auxiliary loss $\mathcal{L}_{aux}$ can be formulated as:

\vspace{-10pt}
\begin{equation}
\mathcal{L}_{aux} = \mathcal{L}_{aux}^{seg} + \mathcal{L}_{aux}^{occupied} 
\end{equation}

The total loss is formulated as:

\vspace{-5pt}
\begin{equation}
\mathcal{L}_{total} = \mathcal{L}_{det} + \mathcal{L}_{occ} + \mathcal{L}_{aux}
\end{equation}

\section{Experiments and Performance Analysis}

\subsection{Dataset and Evaluation Metrics}
\subsubsection{Dataset}


We evaluate our method on OmniHD-Scenes \cite{OmniHDScenes}, VoD \cite{palffy2022multi}, and TJ4DRadSet \cite{zheng2022tj4dradset}. The OmniHD-Scenes features six cameras, six 4D radars, and a 128-beam LiDAR, containing 1501 clips captured in diverse scenarios. The dataset provides annotations for 200 clips, including 3D tracking for four object categories, static scene segmentation, and 11-class semantic occupancy ground truth. The annotated data consists of 11921 keyframes, split into 8321 for training and 3600 for testing.

The VoD dataset covers Delft's campus, suburbs, and old town, with 5139 training and 1296 validation frames. TJ4DRadSet contains 5717 training and 2040 testing frames from diverse road scenarios. Both datasets provide synchronized data from forward-facing 4D radar, camera, and LiDAR, with 3D bounding box annotations.




\subsubsection{Evaluation Metrics}

For the OmniHD-Scenes dataset, we utilize officially defined metrics to evaluate the performance of 3D detection and occupancy prediction within a detection area of ±60m longitudinally and ±40m laterally around the ego vehicle. For 3D detection, we employ mean Average Precision (mAP) along with four mean True Positive metrics (mTP): mean Average Translation Error (mATE), mean Average Scale Error (mASE), mean Average Orientation Error (mAOE), and mean Average Velocity Error (mAVE). Additionally, we adopt the OmniHD-Scenes Detection Score (ODS) to assess comprehensive performance. For 3D occupancy prediction evaluation, we employ two key metrics: mean Intersection over Union (mIoU) for semantic accuracy and scene completion IoU for geometric accuracy. The mIoU is calculated by averaging IoU scores across all semantic categories, where IoU measures the overlap between predicted and ground truth occupancy states for each category.






For both VoD and TJ4DRadSet datasets, we evaluate 3D detection performance using Average Precision (AP) and mean Average Precision (mAP) by following the official evaluation protocols \cite{palffy2022multi, zheng2022tj4dradset}. 

\subsection{Implementation Details}


For the OmniHD-Scenes dataset, we constrain the point cloud range to $(-60, 60)$ m, $(-40, 40)$ m, and $(-3, 5)$ m along the X-, Y-, and Z-axes, respectively. The radar input consists of 3-frame accumulated point clouds, with each point characterized by a feature vector $[x, y, z, Power, SNR, v_{xr}, v_{yr}]$, where $Power$ and $SNR$ represent amplitude and signal-to-noise ratio, while $v_{xr}$ and $v_{yr}$ denote the compensated absolute velocity components. All six camera images are resized to $544 \times 960$. The low-resolution voxel query dimensions $H_V \times W_V \times Z_V$ are set to $80 \times 120 \times 8$, the BEV feature map size $H \times W$ is $160 \times 240$, and the occupancy ground truth resolution is $160 \times 240 \times 16$. We employ a DETR-based detection head with 900 object queries and retain the top 300 predicted boxes with highest confidence scores during inference. For both VoD and TJ4DRadSet datasets, we followed exact same settings as most of the state-of-the-arts \cite{zheng2023rcfusion,xiong2023lxl,SGDet3D}.

Our model implementation is based on the MMDetection3D \cite{contributors2020mmdetection3d} framework and trained using NVIDIA GeForce RTX 4090D GPUs. For the camera encoder, we utilize pre-trained weights from FCOS3D~\cite{wang2021fcos3d} for backbone, maintaining consistency with~\cite{bevformer}. The 4D radar encoder, inherited from RadarPillarNet \cite{zheng2023rcfusion}, is trained from scratch for 3D detection on their respective datasets. We employ the AdamW optimizer for training, with different learning rate for each dataset: for OmniHD-Scenes, we train for 16 epochs with a learning rate of $2\times10^{-4}$; for VoD and TJ4DRadSet, we train for 16 and 20 epochs respectively, both with a learning rate of $1\times10^{-4}$.

All ablation experiments are conducted on the OmniHD-Scenes dataset under a multi-task learning setting with 2 temporal frames, unless otherwise specified. 

\subsection{Performance Comparison with State-of-the-arts}

\begin{table*}[!htbp]
\centering
\caption{Comparison of state-of-the-art approach with ours, for 3D object detection task on the OmniHD-Scenes \emph{test} set. ``C'' denotes camera and ``R'' denotes 4D radar. The last four columns show the AP for each object type. ``Ped.'' and ``LVeh.'' represent pedestrian and large vehicle, respectively. \textbf{Bold} and \underline{underline} denote the first and the second best performances among all the approaches. Doracamom-S does not utilize temporal information. All FPS metrics are tested on a single NVIDIA GeForce RTX 4090D. \textdagger~denotes the end-to-end inference speed for jointly performing both 3D detection and occupancy prediction tasks. 
}
\label{od_performance_omnihd_complete}
\setlength\tabcolsep{2pt}
\resizebox{1.0\textwidth}{!}{
\begin{tabular}{c|c|c|c|cc|cccc|cccc|c}
\hline\noalign{\smallskip}
Methods &   Image Res.& Modality &   Backbone & mAP$\uparrow$ & ODS$\uparrow$ & mATE$\downarrow$ &mASE$\downarrow$ &mAOE$\downarrow$  &mAVE$\downarrow$ & Car$\uparrow$ & Ped.$\uparrow$ & Rider$\uparrow$ & LVeh.$\uparrow$ & FPS$\uparrow$\\
\noalign{\smallskip}
\hline
\noalign{\smallskip}
\hline
\noalign{\smallskip}
PointPillars~\cite{Pointpillars} (CVPR 2019)           & -           & R      & -        & 23.82    & 37.21   &    0.6752    &    0.2447    &       0.3776     &       0.6789     & 52.74 & 0.69 & 28.57 & 13.29 & 62.2\\
RadarPillarNet~\cite{zheng2023rcfusion} (IEEE T-IM 2023)            & -           & R      & -        & 24.88    & 37.81   &  0.6597      &    0.2389    &       0.3736     &  0.6982      & 52.99 & 2.06 & 29.45 & 15.02   & 60.3\\
\noalign{\smallskip}
\hline
\noalign{\smallskip}
LSS-Depth~\cite{bevdepth} (AAAI 2023)              & 544×960       & C      & R50      & 22.44 & 26.01 & 1.0238 & 0.2230    & 0.5942 & 2.0138  & 50.42 & 4.37 & 24.56 & 10.42 & 3.7\\
BEVDet4d~\cite{huang2022bevdet4d} (arXiv 2022)                 &544×960       & C      &R50      & 25.22 & 28.66 & 0.9344 & 0.2352    & 0.5470 & 1.1881  &52.07 &8.07 &31.92 & 8.86 & 18.3\\
BEVFormer-S~\cite{bevformer} (ECCV 2022)       & 544×960        & C       & R50      & 26.49 & 28.10 & 1.1430 & 0.2315    & 0.5799 & 1.6666 & 50.74 & 11.69 & 30.42 & 13.11 & 11.4\\
BEVFormer~\cite{bevformer} (ECCV 2022)             & 544×960        & C       & R50      & 29.17 & 30.54 & 1.1046 & 0.2346    & 0.4889 & 1.0797 & 53.64 & 14.48 & 33.55 & 15.01 & 11.4\\
SparseBEV~\cite{liu2023sparsebev} (ICCV 2023)                 &544×960       & C      &R50      & 30.14 & 30.38 & 1.0272 & 0.2216    & 0.5534 & 1.2188  &54.68 &17.28 &34.52 & 14.08 & 4.9\\
BEVFormer-S~\cite{bevformer} (ECCV 2022)       & 864×1536        & C       & R101-DCN   & 30.10 & 30.55 & 1.0633 & 0.2266    & 0.5331 & 1.6625 & 55.40 & 14.05 & 34.24 & 16.70 & 4.7\\
BEVFormer~\cite{bevformer} (ECCV 2022)             & 864×1536        & C       & R101-DCN   & 32.22 & 32.57 & 1.0637 & 0.2271    & 0.4558 & 1.0683 & 57.61 & 17.37 & 38.02 & 15.87 & 4.7\\
PanoOcc~\cite{wang2024panoocc} (CVPR 2024)                 & 544×960        & C       & R50      & 29.17 & 28.55 & 1.1500 & 0.2446    & 0.6378 & 1.6066  & 51.58 & 15.82 & 35.02 & 14.26 & 5.5$^{\dagger}$\\
BEVFusion~\cite{BEVFusion} (NeurIPS 2022)           & 544×960       & C\&R    & R50      & 33.95 & 43.00 & 0.5730 & \underline{0.2165}    & 0.3814 & 0.7474 & 56.25 & 11.66 & 50.90 & 16.99 &  3.6\\
RCFusion~\cite{zheng2023rcfusion} (IEEE T-IM 2023)                     & 544×960    & C\&R    & R50      & 34.88 & 41.53 & \underline{0.5676} & \textbf{0.2135}    & 0.3711 & 0.9208 & 57.17 & 12.87 & 51.35 & 18.11 &  3.6\\
RCBEVDet~\cite{lin2024rcbevdet} (CVPR 2024)                       &544×960    & C\&R    &R50      & 35.53 & \underline{45.41} & \textbf{0.5138} & 0.2305    & \underline{0.3614} & \underline{0.6825} &\textbf{62.35} &10.11 & \textbf{54.60} & 15.06 &  5.2\\
\noalign{\smallskip}
\hline
\noalign{\smallskip}
\textbf{Doracamom-S~(ours)}                  & 544×960    & C\&R    & R50      &\underline{37.60} & 41.31 & 0.6724 & 0.2329 & 0.4359 & 0.8579 & 58.94  &\underline{17.84} & 52.72 &\underline{20.89} &  4.8$^{\dagger}$\\
\textbf{Doracamom~(ours)}       & 544×960        & C\&R      & R50      &\textbf{39.12}  & \textbf{46.22} & 0.6646 &  0.2331   & \textbf{0.3545} & \textbf{0.6151} & \underline{61.12} & \textbf{19.83} & \underline{53.35} & \textbf{22.18} & 4.2$^{\dagger}$\\
\noalign{\smallskip}
\hline
\end{tabular}}
\end{table*}
\setlength{\tabcolsep}{1.4pt}
\begin{figure*}[htbp]
    \centering
    \includegraphics[width=\linewidth]{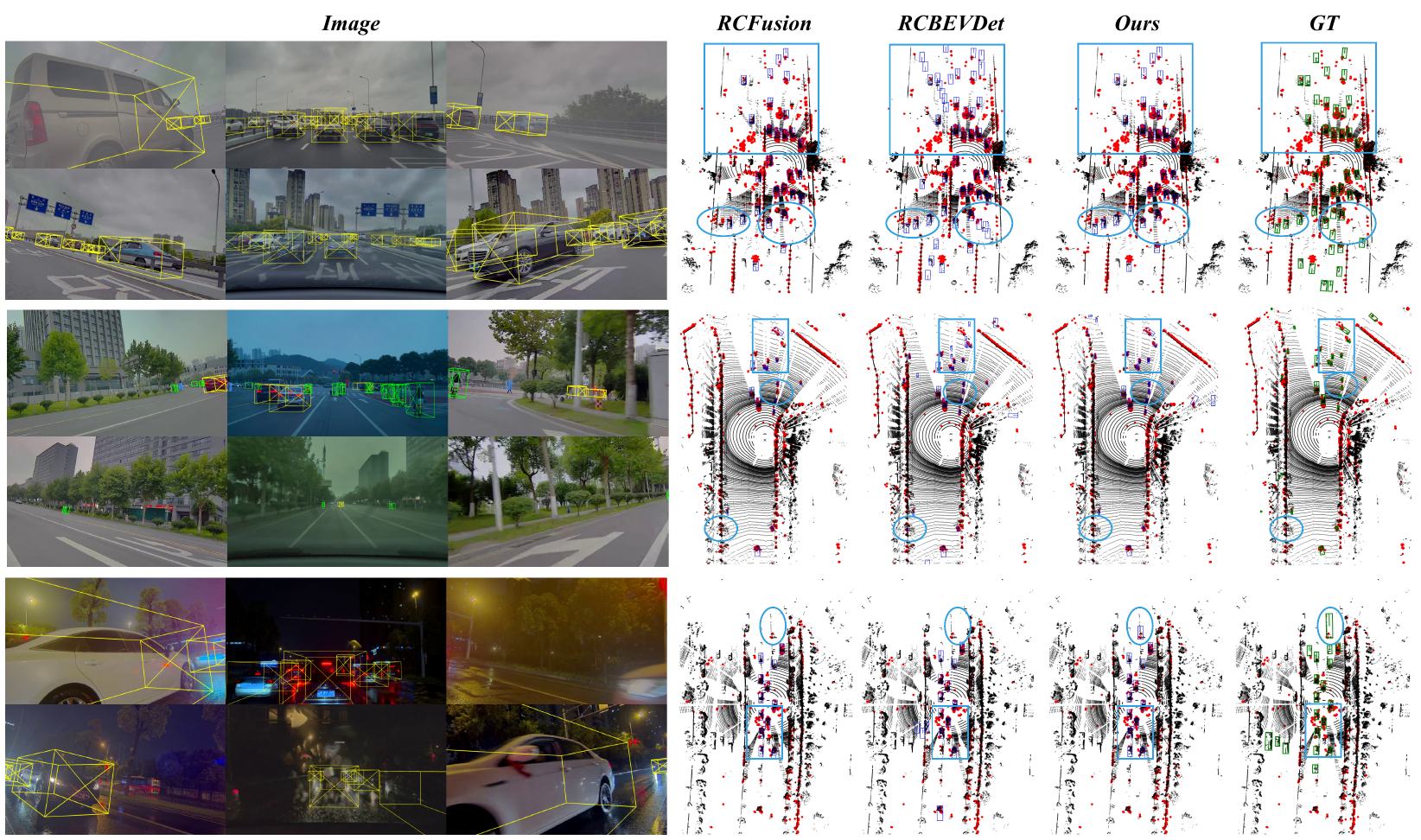}
    \vspace{-0.6cm}
    \caption{Qualitative comparison for 3D object detection on the OmniHD-Scenes test set. The left column shows our 3D bounding box predictions projected onto the six camera views. The right column shows the comparison between RCFusion \cite{zheng2023rcfusion}, RCBEVDet \cite{lin2024rcbevdet}, our method, and the ground truth (GT). In the BEV diagrams, LiDAR points are represented in black, and 4D radar points in \textcolor{Red1}{red}. The highlighted regions indicate areas where our method demonstrates clear advantages in reducing false positives and missed detections compared to the other methods.}
    \label{fig:omnihd_vis_od}
    \vspace{0.2cm}
\end{figure*}


\subsubsection{Results on OmniHD-Scenes}


\textbf{3D Object Detection.} Table~\ref{od_performance_omnihd_complete} presents the performance comparison of different methods on the OmniHD-Scenes test set for 3D object detection. Our proposed Doracamom achieves superior overall performance (39.12 mAP \& 46.22 ODS) compared to other approaches based on 4D radar, camera, or their fusion. Specifically, it surpasses the recent state-of-the-art method RCBEVDet \cite{lin2024rcbevdet} by +3.59 mAP and +0.81 ODS. The performance gap is even larger when compared to other strong baselines like BEVFusion \cite{BEVFusion} (+5.17 mAP, +3.22 ODS) and RCFusion \cite{zheng2023rcfusion} (+4.24 mAP, +4.69 ODS). Even in the single-frame setting, our Doracamom-S outperforms all other methods in terms of mAP. In terms of TP metrics, our method achieves best performance in both mAOE and mAVE, reaching 0.3545 and 0.6151, respectively. Notably, benefiting from the Doppler information inherent in 4D radar, approaches leveraging 4D radar independently or incorporating 4D radar-camera fusion demonstrate remarkable velocity estimation accuracy compared to camera-only methods, exhibiting substantially lower errors in velocity predictions. Furthermore, Doracamom achieves even more precise velocity estimation through efficient exploitation of 4D radar features and the DTE module. For each object category, our method achieves the highest or second-highest AP scores across all classes. Notably, all methods exhibit relatively low detection accuracy for pedestrians and large vehicles, which can be attributed to several challenging factors in the dataset. The dataset contains numerous crowded scenes with dozens of people in small areas, leading to severe occlusions. Moreover, the features extracted from both image and radar data are often incomplete and less distinctive. Additionally, pedestrians occupy only a few grids in the BEV features, further increasing the detection difficulty. For Large Vehicles, their substantial size often results in incomplete contours in both radar and camera views, leading to significant size discrepancies. 

\begin{figure}[t]
    \centering
    \includegraphics[width=\linewidth]{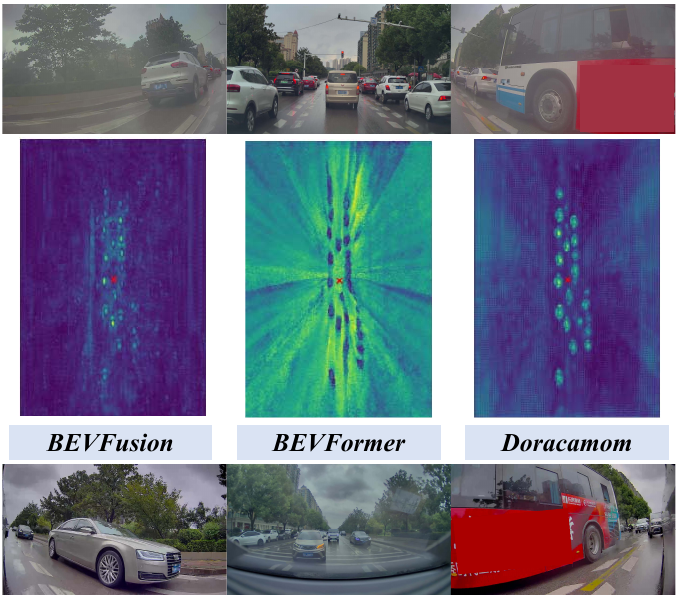}
    \vspace{-0.6cm}
    \caption{Visualization of BEV feature maps from different methods. Compared with BEVFusion \cite{BEVFusion} and BEVFormer \cite{bevformer}, Doracamom generates feature maps with clearer object contours and more prominent features.}
    \label{fig:featuremap}
    \vspace{0.2cm}
\end{figure}

Fig. \ref{fig:omnihd_vis_od} provides a qualitative comparison of Doracamom with other methods on the 3D detection task. The figure shows the performance of our model in various weather conditions, including daytime, nighttime, and rain in complex urban traffic scenes. It is clear that our Doracamom shows superior performance compared to RCFusion and RCBEVDet. The highlighted regions show that our method reduces both missed detections and false positives when dealing with dense objects and distant objects, proving its stronger robustness and generalization ability. Fig. \ref{fig:featuremap} illustrates the BEV feature maps of different methods. It can be observed that the feature map of Doracamom displays distinct object boundaries and highly distinguishable features, with no significant issues such as severe stretching or distortion of the objects.

\definecolor{ncar}{RGB}{255, 165, 0}
\definecolor{npedestrian}{RGB}{128, 0, 128}
\definecolor{nrider}{RGB}{0, 0, 200}
\definecolor{nlarge_vehicle}{RGB}{220, 220, 0}
\definecolor{ncycle}{RGB}{230 ,230 ,230}
\definecolor{nroad_obstacle}{RGB}{255, 69, 0}
\definecolor{ntraffic_fence}{RGB}{0, 0, 150}
\definecolor{ndriveable_surface}{RGB}{135 ,206 ,235}
\definecolor{nsidewalk}{RGB}{200, 200, 200}
\definecolor{nvegetation}{RGB}{34 ,139 ,34}
\definecolor{nmanmade}{RGB}{230, 230, 250}



\begin{table*}[htbp]
\centering
\caption{Comparison of state-of-the-art approach with ours, for 3D occupancy prediction task on the OmniHD-Scenes \emph{test} set. "C" denotes camera and "R" denotes 4D radar. The last eleven columns show the IoU for each semantic type. \textbf{Bold} and \underline{underline} denote the first and the second best performances. Doracamom-S does not utilize temporal information.
}
\label{occ_performance_omnihd_complete}
\begin{threeparttable}
\resizebox{1.0\textwidth}{!}{
\begin{tabular}{c|c|c|c|cc|ccccccccccc}
\hline
\noalign{\smallskip}
Methods     & Image Res. & Modality & Backbone & SC IoU  & mIoU  & \rotatebox{90}{\textcolor{ncar}{$\blacksquare$}car}    & \rotatebox{90}{\textcolor{npedestrian}{$\blacksquare$}pedestrian} & \rotatebox{90}{\textcolor{nrider}{$\blacksquare$}rider} & \rotatebox{90}{\textcolor{nlarge_vehicle}{$\blacksquare$}large vehicle} & \rotatebox{90}{\textcolor{ncycle}{$\blacksquare$}cycle} & \rotatebox{90}{\textcolor{nroad_obstacle}{$\blacksquare$}road obstacle} & \rotatebox{90}{\textcolor{ntraffic_fence}{$\blacksquare$}traffic fence} & \rotatebox{90}{\textcolor{ndriveable_surface}{$\blacksquare$}drive. surf.} & \rotatebox{90}{\textcolor{nsidewalk}{$\blacksquare$}sidewalk} & \rotatebox{90}{\textcolor{nvegetation}{$\blacksquare$}vegetation} & \rotatebox{90}{\textcolor{nmanmade}{$\blacksquare$}manmade} \\
\noalign{\smallskip}
\hline
\noalign{\smallskip}
C-CONet \cite{wang2023openoccupancy} (ICCV 2023)    & 544×960  & C    & R50  & 25.69 & 13.42 & 20.03 & 3.51 & 11.71 & 16.62 & 0.79 & 1.14 & 22.75 & 33.57 & 14.82 & 17.73 & 4.93 \\
SurroundOcc \cite{surroundocc} (ICCV 2023) & 544×960  & C    & R50  & 28.61 & 15.20 & 21.46 & 3.96 & 10.76 & 16.58 & 1.57 & 2.99 & 21.63 & 48.52 & 18.31 & 16.73 & 4.71 \\
BEVFormer-S \cite{bevformer} (ECCV 2022)  & 544×960  & C    & R50  & 27.04 & 14.97 & 20.64 & 5.87 & 14.40 & 16.68 & 1.52 & 3.64 & 20.64 & 46.61 & 16.19 & 14.80 & 3.69 \\
BEVFormer \cite{bevformer} (ECCV 2022) & 544×960  & C    & R50  & 28.42 & 16.23 & 22.73 & 5.45 & 14.70  & 18.21 & 3.09 & 3.87 & 21.54 & 48.15 & 17.58 & 17.77 & 5.48 \\
PanoOcc~\cite{wang2024panoocc} (CVPR 2024)          &544×960       & C      &R50  &26.36  &15.20  &22.42  &5.91    &13.58  & 17.98 & \underline{3.11} & 3.36 & 21.46 & \underline{50.47} & 15.90 & 11.20 & 1.80   \\
\noalign{\smallskip}
\hline
\noalign{\smallskip}
BEVFormer-S \cite{bevformer} (ECCV 2022)  & 864×1536 & C    & R101-DCN & 28.30 & 16.41 & 23.72 & 6.37 & 16.33 & 20.44 & 1.78 & 3.78 & 22.21 & 48.55 & 17.88 & 15.49 & 3.99 \\
BEVFormer \cite{bevformer} (ECCV 2022) & 864×1536 & C    & R101-DCN & 29.74 & 17.49 & 24.90 & 6.48 & 16.45 & 21.49 & 2.87 & 4.62 & 22.51 & 49.92 & \underline{18.59} & 18.53 & 5.96 \\
\noalign{\smallskip}
\hline
\noalign{\smallskip}
BEVFusion~\cite{BEVFusion} (NeurIPS 2022)   & 544×960  & C\&R & R50  & 27.02 & 16.24 & 27.02 & 4.78 & 21.71 & 21.59 & 1.55 & 2.78 & 25.21 & 44.35 & 12.32 & 13.06 & 4.25 \\
M-CONet  \cite{wang2023openoccupancy}  (ICCV 2023)  & 544×960  & C\&R & R50  & 27.74 & 16.08 & 25.21 & 3.42 & 17.53 & 21.46 & 0.88 & 0.58 & \underline{29.88} & 34.48 & 14.89 & \underline{19.57} & \underline{8.98} \\
\noalign{\smallskip}
\hline
\noalign{\smallskip}
\textbf{Doracamom-S~(ours)}            &544×960   & C\&R    &R50  &\underline{31.46} &\underline{19.49} &\underline{30.10} &\underline{6.71} &\underline{23.60} & \underline{24.31} &2.85 &\underline{6.55} & 25.77 &49.72 &16.53 &\underline{19.57} &8.72\\
\textbf{Doracamom~(ours)}     & 544×960    & C\&R     & R50      & \textbf{33.96} & \textbf{21.81} & \textbf{30.81}  & \textbf{7.22}  &  \textbf{24.33}     & \textbf{24.70} &    \textbf{4.49}      & \textbf{7.84} &   \textbf{34.49}        &  \textbf{52.00}         &  \textbf{20.86}            &  \textbf{21.68}   &  \textbf{11.49}    \\

\noalign{\smallskip}
\hline
\end{tabular}}
\end{threeparttable}

\end{table*}
\setlength{\tabcolsep}{1.4pt}

\begin{figure*}[htbp]
    \centering
    \includegraphics[width=\linewidth]{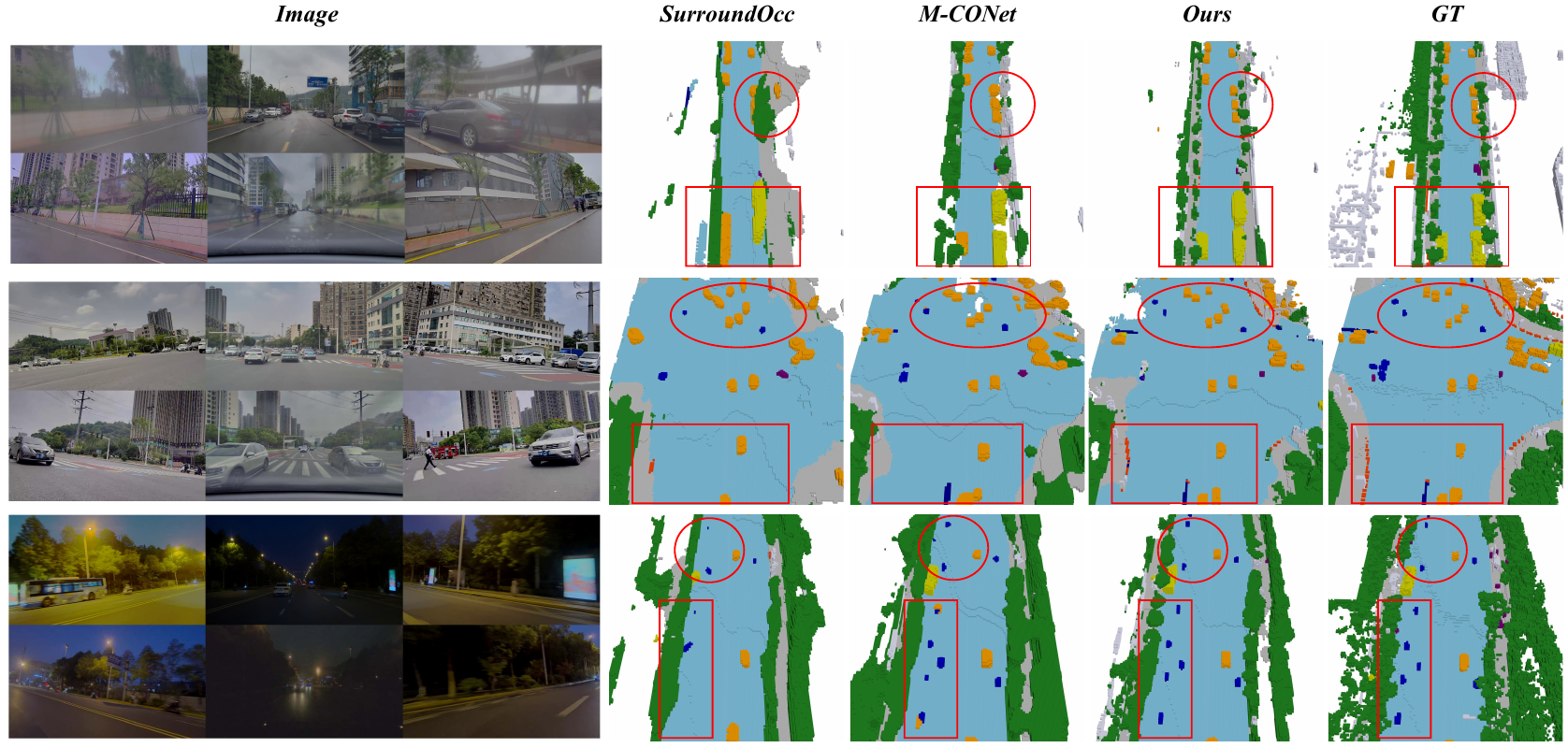}
    \vspace{-0.6cm}
    \caption{Qualitative comparison for semantic occupancy prediction on the OmniHD-Scenes test set. The left column shows the multi-view camera images for reference. The right columns compare the predicted occupancy of SurroundOcc \cite{surroundocc}, M-CONet \cite{wang2023openoccupancy}, our method, and the ground truth (GT), using the same color scheme as Table \ref{occ_performance_omnihd_complete}. The highlighted regions emphasize areas where our method demonstrates superior performance in capturing fine-grained details and achieving higher completeness, especially in challenging scenarios.}
    \label{fig:omnihd_vis_occ}
    \vspace{0.2cm}
\end{figure*}

\begin{table}[t]
\centering
\caption{Comparison of state-of-the-art approach with ours, for 3D object detection task under adverse conditions (night and rainy weather) on the OmniHD-Scenes \emph{test} set. 
}
\label{od_performance_omnihd_adverse_environment}
\resizebox{0.45\textwidth}{!}{
\begin{tabular}{c|c|cccc}
\hline\noalign{\smallskip}
Methods &  Modality & mAP$\uparrow$ & ODS$\uparrow$ & IoU$\uparrow$ &mIoU$\uparrow$  \\
\noalign{\smallskip}
\hline
\noalign{\smallskip}

BEVFormer-S-R101~\cite{bevformer}              & C       &28.11  & 29.36 &26.66   & 15.03     \\
BEVFormer-R101~\cite{bevformer}            & C    & 30.39  & 31.61 &28.41   & 15.84  \\

PanoOcc~\cite{wang2024panoocc}             & C        & 26.09 & 27.16 &24.02  &  14.02      \\
BEVFusion~\cite{BEVFusion}            & C\&R       & 35.83 & 44.95 & 25.32   & 15.36     \\
M-CONet \cite{wang2023openoccupancy}    & C\&R &--- &---  & 26.73 & 15.30  \\
RCBEVDet \cite{lin2024rcbevdet}   & C\&R  &37.49   & \underline{47.32}  &---   &---    \\

\noalign{\smallskip}
\hline
\noalign{\smallskip}
\textbf{Doracamom-S~(ours)}        & C\&R  &\underline{38.75} &43.47 &\underline{29.94} &\underline{18.81}  \\
\textbf{Doracamom~(ours)}            & C\&R      &\textbf{41.86}  & \textbf{48.74} &\textbf{31.06}  &  \textbf{20.30}    \\

\noalign{\smallskip}
\hline
\noalign{\smallskip}
\end{tabular}}

\end{table}
\setlength{\tabcolsep}{1.4pt}
\textbf{3D Semantic Occupancy.} Table \ref{occ_performance_omnihd_complete} presents the performance comparison of different methods on the OmniHD-Scenes test set for occupancy prediction tasks. Our proposed Doracamom achieves superior overall performance (33.96 SC IoU \& 21.81 mIoU) compared to other approaches. When BEVFormer \cite{bevformer} utilizes a larger backbone network (R101-DCN) and higher resolution image input (864×1536), its performance surpasses multi-sensor fusion methods like M-CONet \cite{wang2023openoccupancy} that combines camera and 4D radar data. Nevertheless, with our well-designed architecture, even the Doracamom-S which does not utilize temporal information, that is, without using historical BEV and voxel features, significantly outperforms BEVFormer by +1.72 SC IoU and +2.00 mIoU. Moreover, examining the IoU metrics for each category, we observe that both Doracamom and Doracamom-S achieve superior performance compared to other models in detecting foreground objects of particular interest for object detection, specifically cars, pedestrians, riders, and large vehicles. This performance advantage demonstrates that in a multi-task setting, object detection and occupancy prediction can work synergistically to facilitate both tasks. This advantage is also visually demonstrated in the qualitative comparison in Fig. \ref{fig:omnihd_vis_occ}. The figure compares our method with SurroundOcc, M-CONet, and the ground truth (GT). The results show that Doracamom is superior to the other methods in terms of prediction completeness and capturing fine-grained details in both daytime and nighttime scenarios.

\textbf{Performance under Adverse Conditions.} 
Table \ref{od_performance_omnihd_adverse_environment} denotes the performance of different models under adverse conditions, where Doracamom achieves better results with 41.86 mAP and 48.74 ODS, consistently outperforming other methods and showing stronger robustness. Compared to the results in Table~\ref{od_performance_omnihd_complete}, camera-based methods show decreased detection accuracy, while methods combining 4D radar and camera demonstrate improved performance. Doracamom also achieves 31.06 SC IoU and 20.30 mIoU in occupancy prediction tasks, consistently outperforming other models, which indicates its superior robustness. However, compared to Table \ref{occ_performance_omnihd_complete}, all models show some performance degradation, highlighting the crucial role of camera input in occupancy prediction, as camera performance deterioration directly impacts overall model performance. Nevertheless, with the assistance of 4D radar, the performance degradation remains within an acceptable range. This indicates that 4D radar provides better environmental adaptability in challenging conditions.

\begin{table}[t]
    \belowrulesep=0pt
    \aboverulesep=0pt
    \centering
    \footnotesize
    \caption{Comparison of resources consumption and efficiency with other state-of-the-art fusion methods. (tested on the 4090D GPU).}
    \label{tab:efficiency}
    \renewcommand\arraystretch{1.5}
    \begin{tabular}{c|ccc|cccc}
        \toprule[1.0pt] 
        \multirow{2}{*}{\makebox[1.00cm]{Method}} & 
        \multicolumn{3}{c|}{Resource \& Efficiency} & 
        \multicolumn{4}{c}{Metrics} \\
        \cmidrule(lr){2-4} \cmidrule(lr){5-8}
        & Memory$\downarrow$ & Params.$\downarrow$  & FPS$\uparrow$& mAP$\uparrow$ & ODS$\uparrow$ & IoU$\uparrow$ & mIoU$\uparrow$ \\
        \midrule

        RCFusion \cite{zheng2023rcfusion} &7.96G  &66.02M &3.6 &34.88   &41.53   &--- &---   \\
        BEVFusion-OD \cite{BEVFusion} & 7.96G &57.26M &3.6 & 33.95  & 43.00  &--- &---   \\
        RCBEVDet \cite{lin2024rcbevdet} &\textbf{3.15G}  &85.50M & \textbf{5.2}& 35.53  & \underline{45.41} &--- &---   \\
        BEVFusion-OCC \cite{BEVFusion} & 7.98G &57.22M& 3.2 &--- &---  &  27.02 &16.24    \\
        Doracamom-S & \underline{4.71G}&\textbf{49.63M} &\underline{4.8} &\underline{37.60} &41.31 &\underline{31.46} & \underline{19.49}    \\
        Doracamom & 6.04G & \underline{53.30M}& 4.2 & \textbf{39.12}  & \textbf{46.22}  & \textbf{33.96} & \textbf{21.81} \\
        \bottomrule[1.0pt]
    \end{tabular}
\end{table}

\textbf{Resources Consumption and Efficiency.} 

Table \ref{tab:efficiency} provides a comprehensive comparison of various fusion methods in terms of the resource consumption and inference efficiency. It's crucial to note that the FPS of our Doracamom method represents the speed while performing both 3D object detection and semantic occupancy prediction at the same time. Compared to other methods, the Doracamom series exhibits excellent lightweight performance. Doracamom-S has the lowest parameters (49.63M), and its GPU memory usage is only slightly more than RCBEVDet. In the trade-off between efficiency and performance, Doracamom achieves 4.2 FPS, which is close to the 5.2 FPS of RCBEVDet's single-task operation. However, Doracamom achieves this while handling two tasks simultaneously and delivering SOTA performance of 46.22 ODS and 21.81 mIoU. Compared to BEVFusion, our method has a clear advantage in both speed and accuracy. This highlights that our unified framework finds a better balance between computational efficiency and performance.

As for deployment, the current Doracamom speed is measured under an unoptimized PyTorch framework. However, we're confident that we could significantly improve the model's speed with standard industrial optimizations, like hardware acceleration, quantization, pruning, and resolution compression. Our future work will focus on exploring these optimizations to meet deployment requirements.

\subsubsection{Results on VoD \& TJ4DRadSet}

\begin{table*}[!ht]
\centering
\caption{Comparison of state-of-the-art approach with ours, for 3D object detection task on VoD validation set and TJ4DRadSet test set, respectively. ``C'' denotes camera and ``R'' denotes 4D radar. $\text{\textdagger}$ indicates the use of extra LiDAR data during training procedure. \textbf{Bold} and \underline{underline} denote the first and the second best performances among all the approaches. To ensure fair comparison, Doracamom-S does not utilize temporal information.}
\label{od_performance_vod_tj4dradset}
\resizebox{0.8\textwidth}{!}{
\begin{tabular}{c|c|cccc|ccccc}
\hline
\multirow{2}{*}{Method} & \multirow{2}{*}{Modality} & \multicolumn{4}{c|}{AP$_{3D}$ (\%) of VoD} & \multicolumn{5}{c}{AP$_{3D}$ (\%) of TJ4DRadSet} \\
       &          & Car   & Ped.  & Cyclist  &mAP & Car   & Ped.  & Cyclist & Truck & mAP\\
\hline

RadarPillarNet \cite{zheng2023rcfusion} (IEEE T-IM 2023) & R & 39.30 & 35.10 & 63.63   & 46.01 & 28.45 & 26.24 & 51.57   & 15.20 & 30.37\\
SMURF \cite{liu2023smurf} (IEEE T-IV 2024) & R & 42.31 & 39.09 & 71.50   & 50.97 & 28.47 & 26.22 & 54.61   & 22.64 & 32.99 \\

RCFusion \cite{zheng2023rcfusion} (IEEE T-IM 2023)  & C\&R & 41.70 & 38.95 & 68.31   & 49.65 & 29.72 & \underline{27.17} & \underline{54.93}   & 23.56 & 33.85 \\
RCBEVDet \cite{lin2024rcbevdet} (CVPR 2024) & C\&R & 40.63 & 38.86 & 70.48 & 49.99 & - & - & - & - & - \\
LXL \cite{xiong2023lxl} (IEEE T-IV 2024) & C\&R & 42.33 & 49.48 & \textbf{77.12} & 56.31 & - & - & - & - &36.32\\
RacFormer \cite{chu2025racformer} (CVPR 2025) & C\&R & 47.30 & 46.21 & 69.80 & 54.44 & - & - & - & - & -\\
UniBEVFusion \cite{zhao2024unibevfusion} (IEEE ICRA 2025) & C\&R & 42.22 & 47.11 & 72.94 & 54.09 & 44.26 & 27.92 & 51.11 & 27.75 & 37.76\\
HGSFusion \cite{HGSFusion} (AAAI 2025) & C\&R & 51.67 & \textbf{52.64} & 72.58 & 58.96 & - & - & - & - & 37.21\\
SGDet3D$^{\text{\textdagger}}$ \cite{SGDet3D} (IEEE RA-L 2025) & C\&R & \underline{53.16} & \underline{49.98} & 76.11  & \underline{59.75}  & \textbf{59.43} & 26.57 & 51.30 & \underline{30.00} & \underline{41.82} \\
\hline

\textbf{Doracamom-S~(Ours)} & C\&R &\textbf{53.35}  & 48.94 & \underline{76.99} & \textbf{59.76} & \underline{52.18} &\textbf{33.61}  & \textbf{56.38} & \textbf{34.79}& \textbf{44.24}\\
\hline
\end{tabular}}
\end{table*}

\begin{table}[htp]
    \belowrulesep=0pt
    \aboverulesep=0pt
    \centering
    \footnotesize
    \caption{Ablation study on proposed modules.}
    \label{tab:ablation_main_components}
    \renewcommand\arraystretch{1.1}
\begin{threeparttable}
    \begin{tabular}{c|c|c|c|cccc}
        \toprule[1.0pt] 
        \multirow{2}{*}{\makebox[1.00cm]{Method}} & 
        \multicolumn{3}{c|}{Components} & 
        \multicolumn{4}{c}{Metrics} \\
        \cmidrule(lr){2-4} \cmidrule(lr){5-8}
        & CVQG & DTE & CMF & mAP$\uparrow$ & ODS$\uparrow$ & IoU$\uparrow$ & mIoU$\uparrow$ \\
        \midrule
        Baseline & & & & 35.21 & 38.28 & 29.42 & 18.39 \\
        +CVQG & \checkmark & & & 37.01 & 41.48 & 31.03 & 19.55 \\
        +DTE & \checkmark & \checkmark & & 37.72 & 43.87 & 32.89 & 21.15 \\
        +CMF & \checkmark & \checkmark & \checkmark & \textbf{38.21} & \textbf{44.52} & \textbf{33.16} & \textbf{21.44} \\
        \bottomrule[1.0pt]
    \end{tabular}
\end{threeparttable}
\end{table}

\begin{table}[!htp]
    \belowrulesep=0pt
    \aboverulesep=0pt
    \centering
    \footnotesize
    \caption{Ablation study on CVQG.}
    \label{tab:CVQG}
    \renewcommand\arraystretch{1.5}
\begin{threeparttable}
    \begin{tabular}{c|cc|cccc}
        \toprule[1.0pt]
        Setting  & camera & radar & mAP$\uparrow$ & ODS$\uparrow$ &IoU$\uparrow$ &mIoU$\uparrow$\\
        \midrule
        (a) &--- & --- & 36.92 & 42.94 & 32.51 & 20.30 \\
        (b) & \checkmark & &37.23  &43.27 &32.24 & 20.05  \\
        (c)  & \checkmark &\checkmark & \textbf{38.21} & \textbf{44.52} & \textbf{33.16} & \textbf{21.44} \\

        \bottomrule[1.0pt]
    \end{tabular}
\end{threeparttable}
\end{table}


Table \ref{od_performance_vod_tj4dradset} demonstrates the detection performance of various methods on both VoD and TJ4DRadSet datasets. Since existing state-of-the-art methods do not utilize temporal information, we conducted experiments without temporal data to ensure a fair comparison. On the VoD dataset, Doracamom-S achieves superior detection performance with 59.76 mAP, slightly outperforming the current state-of-the-art method SGDet3D \cite{SGDet3D}, despite the latter utilizing additional LiDAR data for depth supervision. Compared to HGSFusion \cite{HGSFusion} and LXL \cite{xiong2023lxl}, our method shows improvements of 0.8 mAP and 3.45 mAP respectively. In terms of per-category AP metrics, our method achieves the best performance in car detection and second-best performance in cyclist detection.

On the TJ4DRadSet dataset, Doracamom-S demonstrates an even more substantial lead with an mAP of 44.24, surpassing SGDet3D \cite{SGDet3D} and HGSFusion \cite{HGSFusion} by 2.42 mAP and 7.03 mAP respectively. Notably, our method achieves state-of-the-art results across multiple categories including pedestrians, cyclists, and trucks. These results demonstrate that our approach effectively handles both large objects with abundant radar reflection points and smaller objects with fewer reflections, validating the architecture's superior capability in feature fusion and utilization across complex scenarios.

\subsection{Ablation Study}

\textbf{Ablation Study on Main Components.} Table \ref{tab:ablation_main_components} demonstrates the impact of our core components on model performance, including Coarse Voxel Query Generation (CVQG), Dual-branch Temporal Encoder (DTE), and Cross-modal BEV-Voxel Fusion module (CMF). Starting from a baseline model (random initialization, current-frame features, and feature summation), we progressively add components for analysis.

The baseline model achieves 35.21 mAP and 38.28 ODS in detection tasks, while reaching 29.42 IoU and 18.39 mIoU in occupancy prediction. With the introduction of the VQG module, all metrics show significant improvements (+1.80 mAP, +3.20 ODS, +1.61 IoU, +1.16 mIoU), validating its effectiveness in generating high-quality initial queries. Further incorporating the DTE module (2 frames) enhances detection performance to 37.72 mAP and 43.87 ODS, while occupancy prediction improves to 32.89 IoU and 21.15 mIoU, demonstrating its powerful temporal fusion capabilities. Finally, the complete model with the CMF module achieves optimal performance across all metrics, confirming its strong multi-modal feature fusion ability. These results validate the effectiveness of each proposed component and their synergistic benefits when combined within our framework.

\textbf{Ablation Study on CVQG.} Table \ref{tab:CVQG} presents the ablation results of the CVQG module, with random initialization serving as the baseline. When incorporating image semantic features for query initialization, detection performance shows slight improvements (+0.31 mAP, +0.33 ODS), but occupancy prediction performance experiences a minor decrease. Furthermore, when integrating both camera semantic features and 4D radar geometric features for initialization, all metrics demonstrate significant improvements. Compared to the baseline, detection performance increases by 1.29 mAP and 1.58 in ODS, while occupancy prediction improves by 0.65 IoU and 1.14 mIoU. These results demonstrate that CVQG, by effectively combining 4D radar geometric and dynamic features with image semantic priors, successfully enhances the performance in both detection and occupancy prediction tasks.

\begin{table}[t]
    \belowrulesep=0pt
    \aboverulesep=0pt
    \centering
    \footnotesize
    \caption{Ablation study on proposed DTE modules.}
    \label{tab:DTE}
    \renewcommand\arraystretch{1.5}
    \resizebox{0.5\textwidth}{!}{
    \begin{tabular}{c|ccccc|ccccc}
        \toprule[1.0pt] 
        \multirow{2}{*}{\makebox[1.00cm]{Length}} & 
        \multicolumn{5}{c}{Naive ConvBlock} & 
        \multicolumn{5}{c}{Ours} \\
        \cmidrule(lr){2-6} \cmidrule(lr){7-11}
        & mAP$\uparrow$ & ODS$\uparrow$ & IoU$\uparrow$ &mIoU$\uparrow$ &mAVE$\downarrow$& mAP$\uparrow$ & ODS$\uparrow$ & IoU$\uparrow$ & mIoU$\uparrow$&mAVE$\downarrow$ \\
        \midrule
        T = 1 &37.60 &41.31 &31.46 & 19.49 & 0.8579 &37.60 &41.31 &31.46 & 19.49 & 0.8579 \\
        T = 2 &38.13 & 43.96 & 31.85 &20.15 & 0.6927 & 38.21 & 44.52 & 33.16 & 21.44 & 0.6686\\
        T = 3 & 38.81  & 45.05 &32.57&21.03 & 0.6712 &38.75  &45.35  &33.55  &21.64 & 0.6576 \\
        T = 4 &\textbf{39.56}  &45.65  &32.80  &21.51 & 0.6608 & 39.12  & \textbf{46.22}  & \textbf{33.96} & \textbf{21.81} &\textbf{0.6151} \\
        \bottomrule[1.0pt]
    \end{tabular}
    }
\end{table}

\begin{table}[htp]
    \belowrulesep=0pt
    \aboverulesep=0pt
    \centering
    \footnotesize
    \caption{Ablation study on proposed CMF modules.}
    \label{tab:CMF}
    \renewcommand\arraystretch{1.5}
\begin{threeparttable}
    \begin{tabular}{c|cccc}
        \toprule[1.0pt] 
        \multirow{2}{*}{\makebox[1.00cm]{Method}} &  
        \multicolumn{4}{c}{Metrics} \\
        \cmidrule(lr){2-5} 
        & mAP$\uparrow$ & ODS$\uparrow$ & IoU$\uparrow$ & mIoU$\uparrow$ \\
        \midrule
        Add  & 37.72 & 43.87 & 32.59 & 21.11 \\
        Concat  &37.57  &44.17  & 32.94 &20.84  \\
        CMF & \textbf{38.21} & \textbf{44.52} & \textbf{33.16} & \textbf{21.44} \\
        \bottomrule[1.0pt]
    \end{tabular}
\end{threeparttable}
\end{table}

\begin{table}[htp]
    \centering
    \footnotesize
    
    \caption{Ablation study on extrinsic parameter perturbation.}
    \label{tab:extrinsic_perturbation_simple}
    \renewcommand\arraystretch{1.5}
    \begin{threeparttable}
        \begin{tabular}{cc|cccc}
            \toprule[1.0pt]
            \multicolumn{2}{c|}{Max Perturbation} & \multicolumn{4}{c}{Metrics} \\
            \cmidrule(lr){1-2} \cmidrule(lr){3-6}
            Trans.(cm) & Rot. ($^{\circ}$) & mAP$\uparrow$ & ODS$\uparrow$ & IoU$\uparrow$ & mIoU$\uparrow$ \\
            \midrule
            0.0 & 0.0 & \textbf{38.21} & \textbf{44.52} & \textbf{33.16} & \textbf{21.44} \\
            \midrule
            $\pm 2.0$ & $\pm 0.2$ & 38.14 & 44.48 & 32.98 & 21.34 \\
            $\pm 5.0$ & $\pm 0.5$ & 37.80 & 44.23 & 32.23 & 20.87 \\
            $\pm 10.0$ & $\pm 1.0$ & 36.56 & 43.42 & 30.22 & 19.62 \\
            \bottomrule[1.0pt]
        \end{tabular}
    \end{threeparttable}
\end{table}
\textbf{Ablation Study on DTE.} Table \ref{tab:DTE} presents the ablation study results of the DTE module. We use a simple feature concatenation followed by convolution blocks to merge aligned temporal feature sequences as our baseline, termed Naive ConvBlock. The results show that both methods performance improves as the temporal length increases.

When incorporating two-frame temporal information, our proposed DTE module demonstrates stronger performance improvements compared to the naive convolution approach (38.21 mAP vs 38.13 mAP, 44.52 ODS vs 43.96 ODS), with particularly significant gains in occupancy prediction tasks (+1.31 IoU, +1.29 mIoU). With temporal length increased to four frames, although the naive convolution approach achieves optimal mAP, our method obtains the best performance in spatial perception metrics (ODS, IoU, and mIoU). This indicates that the DTE module, through its deformable attention mechanism, can adaptively fuse temporal features, achieving more comprehensive spatio-temporal representations and thus excelling in handling occlusions and enhancing scene understanding. Notably, our method also demonstrates a considerable advantage in velocity estimation (0.6151 vs 0.6608 in mAVE), further validates the effectiveness of the DTE module in spatio-temporal and dynamic object representations.


\textbf{Ablation Study on CMF.} Table \ref{tab:CMF} presents the ablation study results of the CMF module. We compared CMF with two basic feature fusion strategies, feature addition and concatenation. The results demonstrate that our CMF module outperforms all other methods across all metrics. In comparison to the simple addition strategy, our CMF enhances mAP by +0.49 and mIoU by +0.33. Against the concatenation strategy, the benefits of our method are even more evident, with improvements of +0.64 in mAP and +0.60 in mIoU. These results strongly validate the effectiveness of our CMF module design. By combining adaptive attention mechanisms with auxiliary supervision signals, it can more effectively integrate heterogeneous features from the BEV and voxel spaces than simpler fusion methods, thus generating more discriminative and comprehensive feature representations for downstream tasks.

\begin{table}[htbp!]
    \belowrulesep=0pt
    \aboverulesep=0pt
    \centering
    \footnotesize

    \caption{Ablation study on loss components.}
    \label{tab:ablation_loss_components_extended}
    \renewcommand\arraystretch{1.3} 
\begin{threeparttable}
    \begin{tabular}{cccc|cccc}
        \toprule[1.0pt]
        \multicolumn{4}{c|}{Loss Components} & 
        \multicolumn{4}{c}{Metrics} \\
        \cmidrule(lr){1-4} \cmidrule(lr){5-8}
        $\mathcal{L}_{det}$ & $\mathcal{L}_{occ}$ & $\mathcal{L}_{aux}^{seg}$ & $\mathcal{L}_{aux}^{occupied}$ & mAP$\uparrow$ & ODS$\uparrow$ & IoU$\uparrow$ & mIoU$\uparrow$ \\
        \midrule

        \checkmark & & & &37.16 &43.21 &- &- \\

        \checkmark & \checkmark & & &37.84 &44.42 &32.81 &21.30 \\

        \checkmark & \checkmark & \checkmark &  &38.05 &\textbf{44.67} &32.80 &21.35\\

        \checkmark & \checkmark & \checkmark & \checkmark & \textbf{38.21}&44.52 &\textbf{33.16} &\textbf{21.44} \\
        
        \bottomrule[1.0pt]
    \end{tabular}
\end{threeparttable}
\end{table}
\begin{table}[htbp!]
    \belowrulesep=0pt
    \aboverulesep=0pt
    \centering
    \footnotesize

    \caption{Ablation study on encoder layers.}
    \label{tab:encoder_layer}
    \renewcommand\arraystretch{1.5}
\begin{threeparttable}
    \begin{tabular}{c|cccc}
        \toprule[1.0pt] 
        \multirow{2}{*}{\makebox[1.00cm]{Layers}} &  
        \multicolumn{4}{c}{Metrics} \\
        \cmidrule(lr){2-5} 
        & mAP$\uparrow$ & ODS$\uparrow$ & IoU$\uparrow$ & mIoU$\uparrow$ \\
        \midrule
        1  &35.84  &43.52  &32.57  &20.60 \\
        2  & 37.06 &44.34  &32.54  &20.77  \\
        3  & 38.21 & 44.52 & \textbf{33.16} & \textbf{21.44} \\
        4  & \textbf{38.51}  & \textbf{44.55} & 32.59 & 21.27  \\
        \bottomrule[1.0pt]
    \end{tabular}
\end{threeparttable}
\end{table}

\textbf{Ablation study on extrinsic parameter perturbation.}
Table \ref{tab:extrinsic_perturbation_simple} shows the robustness evaluation results of Doracamom when the extrinsic parameters of cameras are disturbed. To simulate real-world calibration drift, we manually added random noise to the translation and rotation parameters of cameras. The noise was sampled independently from a symmetric uniform distribution of $[-\delta, +\delta]$, where $\delta$ is the maximum error. The results clearly show that he performance of Doracamom decreases smoothly as the noise level increases, without any sudden collapse. When a small noise is introduced ($\pm 2.0 cm$, $\pm 0.2^{\circ}$), the performance of our model is almost unaffected, with only 0.04 and 0.1 decrease in ODS and mIoU respectively. Even with larger noise ($\pm 10.0 cm $, $\pm 1.0^{\circ}$), the model still maintained a relatively high performance level, with 43.42 ODS and 19.62 mIoU. This demonstrates the robustness of Doracamom against a certain range of calibration errors, which is crucial for practical deployment. In the future, we can make the model more robust by training with calibration noise or by adding a lightweight online self-calibration module.

\textbf{Ablation study on loss components.} Table \ref{tab:ablation_loss_components_extended} shows the impact of different loss functions on the model performance. We first use a single-task model that only uses $\mathcal{L}_{det}$ for 3D object detection as the baseline, with a performance of 37.16 mAP and 43.21 ODS. When we add the $\mathcal{L}_{occ}$ to this baseline for multi-task training, mAP and ODS increase by 0.68 and 1.21, respectively. This demonstrates that adding the occupancy prediction task helps Doracamom learn richer spatial features, which positively affects the detection task. When we next add the BEV segmentation auxiliary loss $\mathcal{L}_{aux}^{seg}$, nearly all metrics improve. This indicates that guiding the model to better distinguish between foreground and background in the BEV space is beneficial for both 3D semantic understanding and object detection. when we further add the $\mathcal{L}_{aux}^{occupied}$, Doracamom achieves its highest scores for mAP, IoU, and mIoU. This shows that combining the two auxiliary losses guides the model to learn more comprehensive features, resulting in the best overall performance across both tasks.

\textbf{Ablation study of encoder layers.}
Table \ref{tab:encoder_layer} shows ablation study results on the number of Transformer layers $L$ in the Voxel Queries Encoder. We observe that as $L$ increased from 1 to 3, most of the metrics show significant and consistent improvement. For instance, mAP increases from 35.84 to 38.21 and mIoU reaches at 21.44. This proves that a deeper encoder can effectively learn more fine-grained features. However, when $L$ increases to 4, the performance gain becomes very limited, and the metrics for occupancy prediction actually decrease. This indicates that the benefits of adding more depth are diminishing. Therefore, considering the trade-off between performance and computational efficiency, we choose $L=3$ as the final configuration for our model.

\section{Conclusion and Future Work}

In this paper, we present Doracamom, the first unified multi-task perception framework with multi-view camera and 4D radar fusion. Specifically, we propose a coarse voxel query generator that initializes geometrically and semantically aware voxel queries to effectively leverage multi-modal features; a dual-branch temporal encoder that adaptively fuses temporal features in both BEV and voxel spaces while considering dynamic and static elements in the environment; and finally, an attention-based cross-modal BEV-voxel fusion module that adaptively integrates heterogeneous features and incorporates auxiliary tasks to address feature ambiguity, generating high-quality feature representations for downstream tasks. Experimental results on three datasets, OmniHD-Scenes, VoD, and TJ4DRadSet, demonstrate that our method achieves state-of-the-art performance in both 3D object detection and 3D semantic occupancy prediction tasks.

In the future, we will explore more robust temporal modeling mechanisms, such as introducing explicit motion prediction to guide feature alignment or integrating instance-level object tracking to further address the challenge of feature association in dense dynamic scenes. Furthermore, we will focus on optimizing the model for deployment by applying engineering practices such as hardware acceleration, model pruning, and quantization to meet real-time  requirements.

\footnotesize
\bibliographystyle{IEEEtran}
\bibliography{refs}

\end{document}